\documentclass{article}

\usepackage{PRIMEarxiv}

\usepackage[utf8]{inputenc} 
\usepackage[T1]{fontenc}    
\usepackage{hyperref}       
\usepackage{url}            
\usepackage{booktabs}       
\usepackage{amsfonts}       
\usepackage{nicefrac}       
\usepackage{microtype}      
\usepackage{lipsum}
\usepackage{fancyhdr}       
\usepackage{graphicx}       
\graphicspath{{media/}}     

\usepackage{natbib}
\usepackage{soul}
\usepackage{amsmath, enumitem}
\usepackage{bm}
\usepackage{booktabs}
\usepackage{multirow}

\newcommand{\algo}{\textsc{DPEpiNN}}

\newcommand{\xsb}{x^{pub}}
\newcommand{\xsbbf}{\mathbf{x}^{pub}}
\newcommand{\xpriv}{x^{priv}}
\newcommand{\xprivbf}{\mathbf{x}^{priv}}

\newcommand{\Minf}{M_{inf}}
\usepackage{hyperref}
\newcommand{\Dtran}{D^{Tran}}
\newcommand{\parameterNN}{\textsc{parameterNN}}
\newcommand{\Dcal}{\mathcal{D}}

\newtheorem{dfn}{Definition}

\title{Improving Epidemic Analyses with Privacy-Preserving Integration of Sensitive Data}

\author{
  Zihan Guan \\
  Department of Computer Science \\
  University of Virginia \\
  Charlottesville, VA, USA \\
  \texttt{bxv6gs@virginia.edu} \\
  \And
  Zhiyuan Zhao \\
  College of Computing \\
  Georgia Institute of Technology \\
  Atlanta, GA, USA \\
  \texttt{leozhao1997@gatech.edu} \\
  \And
  Fengwei Tian \\
  Department of ECE \\
  University of Arizona \\
  Tucson, AZ, USA \\
  \texttt{fengtian@arizona.edu} \\
  \And
  Dung Nguyen \\
  Department of Computer Science \\
  University of Virginia \\
  Charlottesville, VA, USA \\
  \texttt{dungn@virginia.edu} \\
  \And
  Payel Bhattacharjee \\
  Department of ECE \\
  University of Arizona \\
  Tucson, AZ, USA \\
  \texttt{payelb@arizona.edu} \\
  \And
  Ravi Tandon \\
  Department of ECE \\
  University of Arizona \\
  Tucson, AZ, USA \\
  \texttt{tandonr@arizona.edu} \\
  \And
  B. Aditya Prakash \\
  College of Computing \\
  Georgia Institute of Technology \\
  Atlanta, GA, USA \\
  \texttt{badityap@cc.gatech.edu} \\
  \And
  Anil Vullikanti \\
  Department of Computer Science \\
  University of Virginia \\
  Charlottesville, VA, USA \\
  \texttt{vsakumar@virginia.edu} \\
}

\begin{document}
\maketitle

\begin{abstract}
Epidemic analyses increasingly rely on heterogeneous datasets, many of which are sensitive and require strong privacy protection. Although differential privacy (DP) has become a standard in machine learning and data sharing, its adoption in epidemiological modeling remains limited.
In this work, we introduce DPEpiNN, a unified framework that integrates deep neural networks with a mechanistic SEIRM-based metapopulation model under formal DP guarantees. DPEpiNN supports multiple epidemic tasks (including multi-step forecasting, nowcasting, effective reproduction number $(R_t)$ estimation, and intervention analysis) within a single differentiable pipeline. 
The framework jointly learns epidemic parameters from heterogeneous public and sensitive datasets, while ensuring privacy via input perturbation mechanisms.
We evaluate DPEpiNN using COVID-19 data from three regions. Results show that incorporating sensitive datasets substantially improves predictive performance even under strong privacy constraints. Compared with a deep learning baseline, DPEpiNN achieves higher accuracy in forecasting and nowcasting while producing reliable estimates of $R_t$.
Furthermore, the learned epidemic transmission models remain inherently private due to the post-processing property of differential privacy, enabling downstream policy analyses such as simulation of social distancing interventions.
Our work demonstrates that interpretability (through mechanistic modeling), predictive accuracy (through neural integration), and rigorous privacy guarantees can be jointly achieved in modern epidemic modeling.
\end{abstract}

\keywords{Epidemics Analysis, Differential Privacy, Deep Learning}

\section{Introduction}
\label{sec:intro} 

\begin{figure*}[!t]
    \centering
\includegraphics[width=\linewidth]{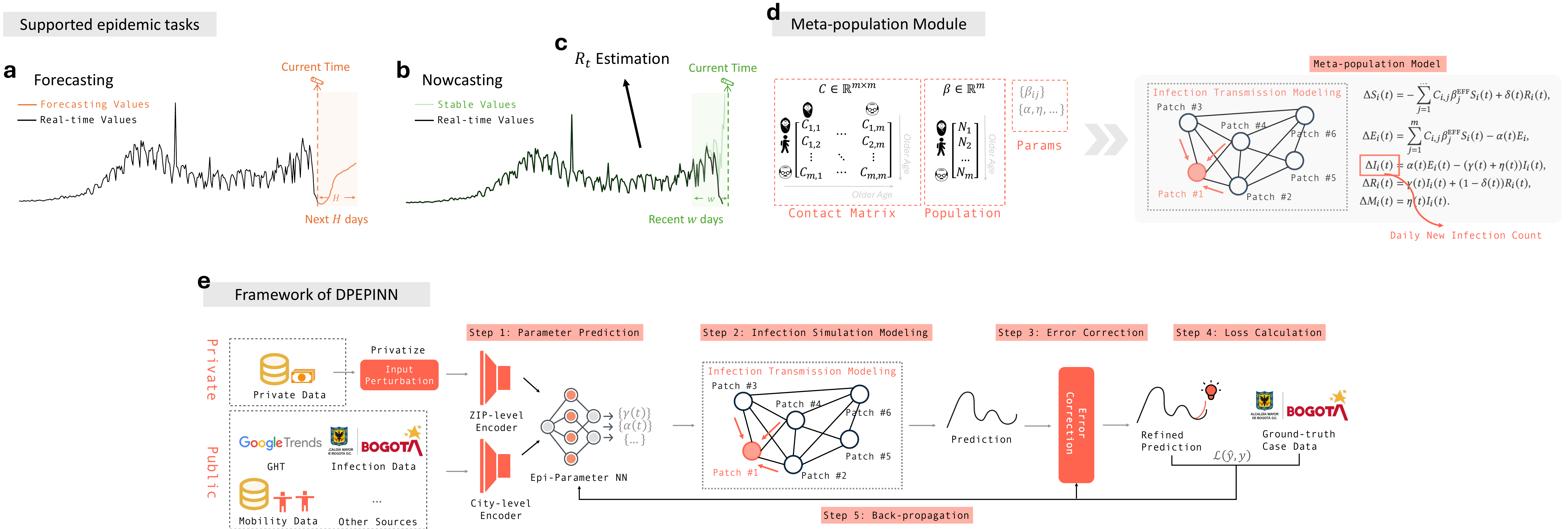}
\caption{\textbf{Overview of epidemic analyses supported by \algo{}.} 
(a) The goal of forecasting is to predict the targets for the next $H$ days given the time series observed at current time $T$;
(b) Nowcasting involves revising the currently observed time series to its stable versions. 
(c) This is used to estimate the effective reproduction number. $R_t$;
(d) The meta-population module takes a public contact matrix $C$ across age groups, a public population vector $\beta$ stratified by age, and the predicted epidemic parameters as input. Using the age-stratified pandemic transmission equations, the meta-population model simulates and aggregates the daily new infection count (i.e., the prediction target) for each time stamp $t$.
(e) \algo{} consists of three modules: \parameterNN{}, Meta-population Model, and Error Correction Adapter. Datasets are processed by the \parameterNN{} modules to generate epidemic parameters using encoders of varying granularities (Step 1). These parameters are then input into the meta-population model to generate pandemic simulations (Step 2). An error correction adapter further corrects errors in the simulations adaptively (Step 3). 
After loss computation (Step 4), the gradients are back-propagated to the \parameterNN{} to further update the network (Step 5). 
The input module privatizes the sensitive time series for a given privacy budget $(\epsilon, \delta)$,
by using input perturbation method (e.g., randomized response and Laplace mechanism).  
}
\label{fig:overview_two_in_one}
\end{figure*}

Epidemic analyses span a very wide range of questions, such as forecasting and nowcasting at different kinds of spatio-temporal resolutions, disease surveillance and inference, estimation of epidemic parameters, analysis of different kinds of interventions and behavioral responses and scenario projections, e.g., \cite{marathe:cacm13, chopra2022differentiable, kamarthi2021back2future}; see Figure \ref{fig:overview_two_in_one}(a-c) for illustrations of these questions.
A broad class of (data-driven) statistical and machine learning methods, e.g., ARIMA and time series, and deep learning techniques,
have been used for forecasting problems.
Reported data are often incomplete and delayed due to various reasons, and subject to retrospective corrections, e.g., \cite{kamarthi2021back2future}; 
nowcasting involves learning such error patterns to improve estimates of current cases.
For questions such as evaluating interventions and scenario projections, 
mechanistic models of epidemic transmission (such as agent-based and metapopulation models, e.g.,~\cite{adiga2022enhancing, mistry2021inferring,venkatramanan2019optimizing}) are commonly used. 

Diverse kinds of datasets are used in forecasting and modeling tasks, such as 
epidemiological datasets such as line lists, e.g.,~\cite{adiga2020data, xu2020epidemiological, COVID19C29:online},
social media, e.g,~\cite{culotta2010towards, abouzahra2021twitter, chen2014flu},
online search and website logs, e.g.,~\cite{ginsberg2009detecting, mciver2014wikipedia, polgreen2007use}, 
symptomatic online surveys, e.g.,~\cite{smolinski2015flu, salomon2021us},
medical and wellness devices, e.g.,~\cite{miller2018smartphone, leuba2020tracking},
retail and commerce data~\cite{nsoesie2014guess, miliou2021predicting}, detailed mobility traces~\cite{pepe2020covid, klise2021analysis}, detailed disease surveillance, behavioral surveys, and
financial data~\cite{mulay2020pandemic}.
Some of these are non-traditional datasets and have been found very useful in epidemic analyses.
However, many of these datasets contain sensitive information that easily leaks individuals' privacy if used without adequate protection.
For instance, it was shown that data released by the US Census was vulnerable to privacy attacks \cite{abowd20232010}.
Networked data has also been shown to be vulnerable to different kinds of privacy attacks, which can reconstruct some sensitive and private information, such as node and edge attributes, e.g.,
\cite{wu2022linkteller, gong2018attribute, zari2024node}.
This has motivated a lot of interest in providing privacy protections for sensitive data.
Among various privacy models, differential privacy (DP) has emerged as the \textit{de facto} standard notion for privacy~\cite{10.1145/1065167.1065184, dwork2011differential, dwork:fttcs14, nissim:stoc07}, since it makes no assumptions of the capabilities of an adversary.
Sophisticated DP techniques have been developed for a broad class of machine learning methods, using diverse kinds of datasets, e.g., \cite{ponomareva2023dp, ji2014differential, blanco2022critical}.
DP is now used by US Census \cite{abowd2018us}, has been proposed for satisfying the General Data Protection Regulation (GDPR) requirements~\cite{cummings2018role} and used by industry~\cite{analyticsexposure, team2017learning}. 
Yet, the adoption of DP in epidemiology and public health has been limited. 
Recent initiatives, such as the U.S.–U.K. Privacy Enhancing Technologies (PETs)  Challenge for Public Health~\cite{nsf-pet, harrison:pet-challenge} and the “PETs for Public Health Challenge” organized by data.org~\cite{pet-challenge}, have underscored this gap, calling for frameworks that can integrate differential privacy mechanisms directly into pipelines for epidemic analyses without compromising model accuracy or interpretability.

This gap leads to our central research problem: 
\textit{How can we design techniques for epidemic analyses, such as forecasting and learning mechanistic models of transmission dynamics from heterogeneous and sensitive datasets while simultaneously ensuring DP guarantees}?
Epidemic analyses pose several novel challenges for DP, and new advances are needed to provide privacy protections for such problems.
Public and private datasets provide complementary but fundamentally different information for epidemic analysis, are highly heterogeneous and often misaligned. 
Public data sources (e.g., reported case counts, hospitalizations, wastewater surveillance, mobility aggregates) offer broad population-level coverage and long temporal histories, enabling robust estimation of global trends and transmission regimes. In contrast, private datasets (e.g., clinical records, financial transactions, fine-grained mobility traces) contain high-resolution signals about individual behavior, contact patterns, and intervention responses that are critical for identifying latent mechanisms, early warning indicators, and heterogeneous effects that are invisible in aggregate data. 
Leveraging both is therefore essential for accurate forecasting and mechanistic understanding.
The heterogeneity and misalignment in sampling bias, spatial granularity, noise characteristics, and semantic meaning
make joint modeling difficult, even without privacy constraints.

To this end, we introduce \algo{} (Differentially Private Epidemic Model with Neural Networks), a unified framework designed to support diverse epidemic analyses tasks under rigorous differential privacy (DP) guarantees. 
Specifically, we use \algo{} for forecasting (predicting future trends), nowcasting (correcting real-time reported values) and estimation of the effective reproduction number ($R_t$), as well as for learning an SEIRM-type disease model and evaluating interventions (Figures \ref{fig:overview_two_in_one} (a-c).
\algo{} integrates heterogeneous datasets (this includes public datasets signals such as mortality, mobility and search trends, as well as sensitive datasets such as financial transactions) within a unified differentiable pipeline, which integrates neural networks and an SEIRM metapopulation model, as shown in Figure \ref{fig:overview_two_in_one}(d).

\algo{} consists of three modules (Figure \ref{fig:overview_two_in_one}(e)): \parameterNN{}, Meta-population Model, and Error Correction Adapter, which are trained \emph{jointly}.
\parameterNN{} is a neural network which processes all the heterogeneous inputs (including the private data, as described later) using multiple encoders, and learns the parameters of an SEIRM meta-population model.
The Adapter is another neural network, which takes the projections from the SEIRM model and learns how to  corrects it for forecasting and nowcasting.
The entire algorithm is trained jointly: 
The loss is computed by comparing the output of the Adapter (depending on whether it is forecasting or nowcasting) with the ground truth infection data, and the gradients are back-propagated to update the weights of all the neural networks.
Sensitive datasets are provided as inputs to \parameterNN{} after input perturbation techniques (discussed later), which ensure privacy of the entire algorithm.

Thus, by integrating deep neural networks with the SEIRM disease model, \algo{} learns high-resolution and interpretable transmission dynamics from heterogeneous data sources, which are protected by the post-processing property of DP---this means that the SEIRM model can be used for epidemic analyses, such as evaluation of interventions.
This is a significant difference from methods considered in prior literature for learning epidemic models, which either calibrate model parameters using purely statistical techniques, e.g., \cite{kerr2021covasim, marathe:cacm13, adiga2020mathematical};
in particular, prior methods cannot be used when the sensitive data cannot be directly mapped to epidemic model components, which is the case here.
We illustrate the effectiveness of \algo{} by showing results during the COVID pandemic for three regions, and using two different types of datasets viewed as being sensitive with user level privacy in one of them and event level privacy in the other.
We find that these sensitive datasets lead to improved performance, even with privacy protections. 

\section{Results}
\begin{figure*}[!t]
    \centering
    \includegraphics[width=\linewidth]{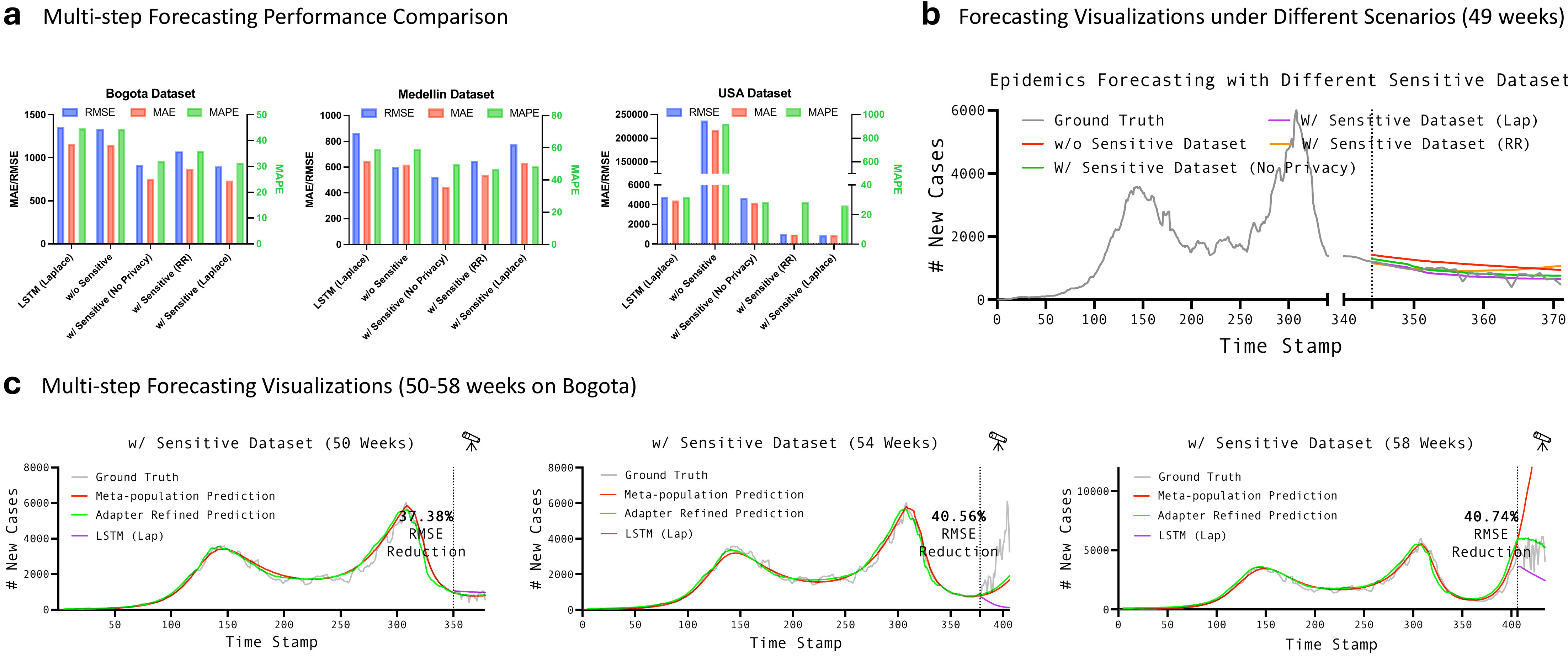}
    \caption{\textbf{Forecasting Performance a.}
Multi-step forecasting performance on the Bogota, Medellin, and the USA datasets. The privacy budget is $\epsilon=1$. \textbf{b.} Visualizations of single-step forecasting on the Bogotá dataset. Different colors indicate different settings: red denotes forecasting without sensitive data; green denotes forecasting with sensitive data under no privacy protection; purple denotes forecasting with sensitive data privatized using the Laplace mechanism; and orange denotes forecasting with sensitive data privatized using the RR mechanism. \textbf{c.} 
Visualizations of the multi-step forecasting, where the training periods shift from 50 weeks to 58 weeks and the testing period is fixed as 4-weeks (28 days). Red and green denote forecasting from the meta-population module and the error correction adapter, respectively. Purple denotes forecasting from the LSTM model.}
    \label{fig:forecasting_results}
\end{figure*}

\subsection{Evaluation Setup for the \algo{}}
\paragraph*{Datasets}

We illustrate the effectiveness of \algo{} by showing results for three regions, the Colombian cities of Bogot\'a and Medell\'in, and the United States, during the COVID pandemic.
For all these regions, we use COVID infection cases and mortality, Google Health Trends, and Google Mobility as public datasets.
For Bogot\'a and Medell\'in, we also use a realistic but synthetic financial dataset consisting of merchant-level credit card transaction data released by data.org as part of a ``PETs for public health challenge''~\cite{pet-challenge}; each entry in the data records the zip code, category, time, transaction amount, location, and merchant id.
We consider this as a sensitive dataset, and refer to it as the ``transaction dataset'' (More details are in the Appendix).
For the US, we treat hospitalization information (e.g., “Hospitalized Increase Count”) released by the U.S. Department of Health \& Human Services as private.
Complete details of these datasets are presented in the Appendix.

\paragraph*{Privacy Model} We say that a (randomized) algorithm $M:\Dcal\to R$ satisfies  $(\epsilon, \delta)$-DP if for all subsets $S\subset R$, and for all $D, D'\in \Dcal$ with $D\sim D'$, we have $Pr[M(D) \in S] \leq e^{\epsilon} Pr[M(D') \in S] + \delta$~\cite{dwork:fttcs14}.
Simply put, a DP mechanism ensures that the result changes in a controlled manner if the dataset is changed from $D$ to its neighbor $D'$.
For the transaction dataset, we consider user-level privacy, where users correspond to merchants (this was a requirement in the PET challenge~\cite{pet-challenge}); here, $D, D'$ differ in the entire data associated with a single merchant.
For the USA dataset, we consider an event-level privacy model, where $D, D'$ differ in the data associated with a single timestamp.
These two kinds of privacy models are considered in the privacy literature~\cite{dwork2010differential, ghazi2023user}.



\paragraph*{Evaluation Metrics} For performance evaluation in all of our experiments, we use three popular metrics, including root mean square error (RMSE), mean absolute error (MAE), and mean absolute percentage error (MAPE). For any given prediction $\bm{y}'_t$ and the corresponding ground-truth values $\bm{y}_t$ for an interval $[t_1, t_2]$, the RMSE for this interval is formally written as $\sqrt{\frac{1}{t_2-t_1+1}\sum_{t=t_1}^{t_2} (\bm{y}_t-\bm{y}_t')^2}$, the MAE is $\frac{1}{t_2-t_1+1}\sum_{t=t_1}^{t_2} |\bm{y}_t-\bm{y}_t'|$, and the MAPE is $\frac{1}{t_2-t_1+1}\sum_{t=t_1}^{t_2} |\frac{\bm{y}_t-\bm{y}_t'}{\bm{y}_t}|$.

\subsection{Forecasting Performance}
Here, we consider the forecasting of ``number of infected cases''.
For all of the experiments, we split the dataset into training period $[1, T]$ and testing period $[T+1, T+H]$, where $H$ is the forecast horizon (taken to be 28 days). 
The training period is used for training the parameter models, and the testing period is used to evaluate the performance of our method. In other words, this means that we aim to employ \algo{} to predict daily new infected case numbers.


\paragraph*{Multi-step Evaluations}
We evaluate the forecasting performance of \algo{} under varying training window lengths $T$ and report the average performance metrics. Specifically, we train \algo{} using 49, 50, 54, and 58 weeks of data from the Bogota and Medellín datasets, and 40, 41, 44, and 49 weeks of data from the USA dataset. We report the average performance metrics across these four runs. Figure~\ref{fig:forecasting_results}(a) reports the average performance across these settings. As shown, \algo{} consistently outperforms the baseline LSTM method. Moreover, we find that for all the datasets, the sensitive data improves performance.
For instance, the RMSE for Bogot\'a, Medell\'in and the USA dataset, compared to LSTM, is lower by over 33.67\%, 10.34\% and 81.30\%, respectively, when the sensitive data are used. This highlights the power of integration of deep learning and disease transmission models in \algo{}.

To further illustrate the performance difference, we visualize the forecasting results on the Bogota dataset. In Figure~\ref{fig:forecasting_results}(b), we visualize the forecasting results under different scenarios: w/o sensitive data, w/ sensitive data (No Privacy), w/ sensitive data privatized by the Laplace mechanism, and w/o sensitive data privatized by the RR mechanism. 
It is clearly shown that the forecasting trajectory deviates much from the ground truth values without the sensitive transaction dataset. In Figure~\ref{fig:forecasting_results}(c), we compare the predictions of our method—using sensitive data privatized via the Laplacian mechanism ($\epsilon = 1$) with those of the LSTM baseline. \ul{Our method more accurately captures the underlying trends of the time series}. For example, at weeks 54 and 58, the LSTM's predictions substantially deviate from the observed case curve, whereas our method closely follows the true trend, showing over 37\% lower RMSE values. Additionally, \ul{the refined predictions generated by the adapter are consistently better than those produced by the meta-population model alone}, validating our intuition of using the adapter as a local error-correction module. Specifically, the RMSE values of the adapter forecasting outputs show an average reduction of 30\%.

\paragraph*{Impact of privacy budget $\epsilon$}
Privacy is more restrictive for smaller values of $\epsilon$. 
We report the performance of our method under different privacy budgets $\epsilon$ with the Laplace mechanism in Table~\ref{tab:impact_eps} in the Appendix. 
\algo{} generally has good performance, 
outperforming the LSTM baseline ($\epsilon=1$) even with $\epsilon=10$. 
We observe a good privacy accuracy tradeoff, with error metrics generally decreasing as $\epsilon$ increases; however, the exact relationship remains unclear and warrants further investigation. This observation aligns with the general intuition that a more relaxed privacy budget introduces less noise in the input perturbation module, allowing the underlying patterns in the private features to be better preserved.


\begin{figure*}
    \centering
    \includegraphics[width=\linewidth]{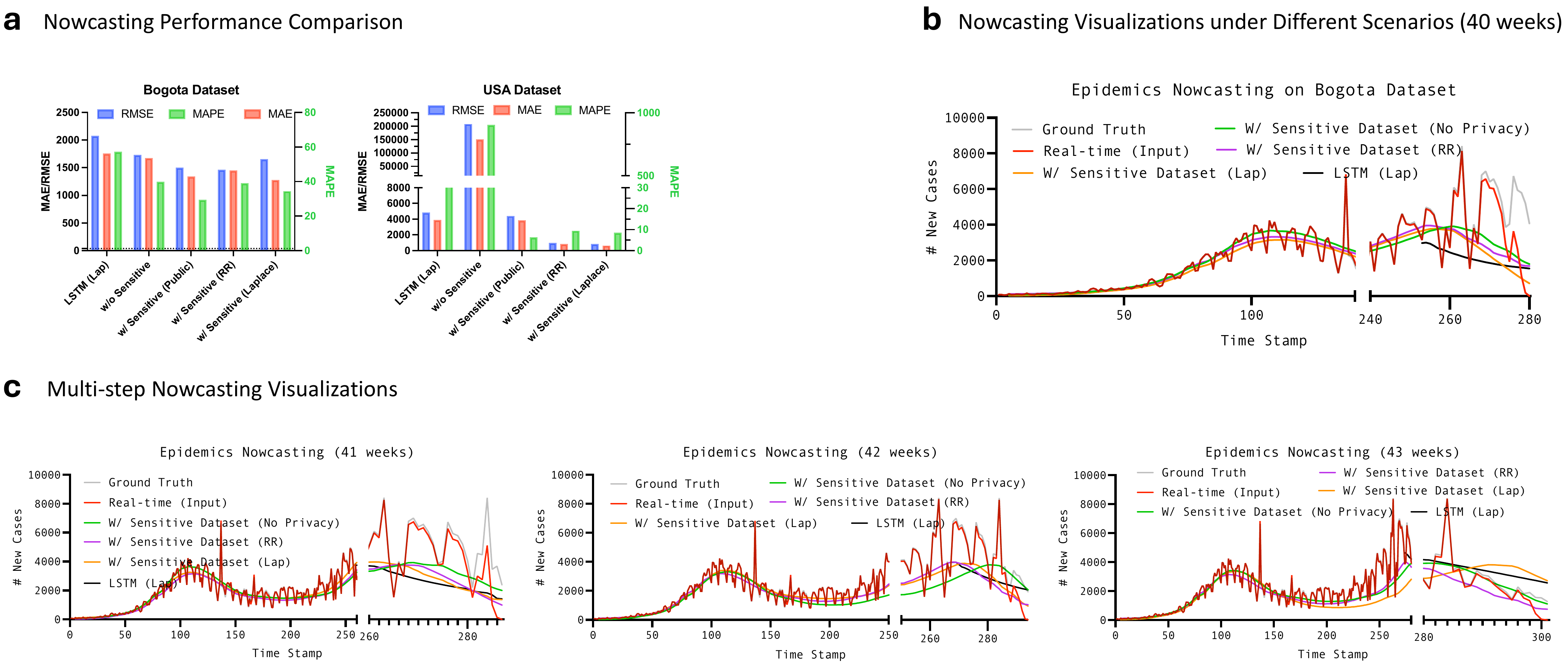}
    \caption{\textbf{Nowcasting Performance a.} The left panel shows the single-step nowcasting performance on the Bogota and the USA datasets. The right panel shows the multi-step nowcasting performance on the Bogota and the USA datasets. \textbf{b.} Visualizations of single-step nowcasting results. Different colors indicate different settings: red denotes the real time input; green denotes nowcasting with sensitive data under no privacy protection; purple denotes nowcasting with sensitive data privatized using the RR mechanism; and orange denotes nowcasting with sensitive data privatized using the Laplace mechanism. \textbf{c.} Visualizations of multi-step nowcasting results. The model is trained on a 40-week period and then continuously deployed using shifting window sizes of 41, 42, and 43 weeks. Red denotes the real-time input; green denotes nowcasts using sensitive data under no privacy protection; purple denotes nowcasts using sensitive data privatized with the RR mechanism; and orange denotes nowcasts using sensitive data privatized with the Laplace mechanism.}
    \label{fig:nowcasting_results}
\end{figure*}

\subsection{Nowcasting Performance} 
Due to reporting delays and systematic errors arising from factors such as testing backlogs and data revisions, epidemiological surveillance data are inherently incomplete and subject to retrospective corrections~\cite{kamarthi2021back2future, choi2012predicting, gunther2021nowcasting}.
The goal of nowcasting is to infer the final, stable case counts from temporarily observed, provisional reports. Formally, let $\{\mathbf{y}^{(T)}_t\}_{t=1}^T$ denote the time series observed up to time $T$, where $\mathbf{y}_t^{(T)}$ represents the reported value at time $t$ as of reporting time $T$. The objective of nowcasting is to revise the most recent observations $\textbf{y}^{(T)}_{T},...,\textbf{y}^{(T)}_{T-w}$ so that they approximate their eventual stable values $\textbf{y}^{(t_f)}_{T},...,\textbf{y}^{(t_f)}_{T-w}$, where $t_f$ denotes the time by which reported data have stabilized.
This is defined formally in (Appendix~\ref{sec:supplementary}).

\algo{} can be readily adapted for this task by using the historical real-time sequences as input and the corresponding stable values as ground truth during training. 
During inference, the current real-time sequence is fed into the model to obtain the refined values (see `Nowcasting Experimental Setups' in the Supplementary File for more details).


\paragraph*{Multi-step Evaluations} We evaluate the nowcasting performance of \algo{} under different training window lengths $T$ and report the averaged metrics. Specifically, we train \algo{} using 40, 41, 42, and 43 weeks of data from both the Bogotá and USA datasets, and then average the performance across these four runs. As shown in Figure~\ref{fig:nowcasting_results}(a), \algo{} consistently outperforms the LSTM baseline on both datasets. For example, when private datasets are used, the RMSE for Bogotá and the USA is reduced by more than 20.45\% and 81.9\%, respectively, compared with LSTM.

To further illustrate the nowcasting behavior, we visualize the results on the Bogota dataset. In Figure~\ref{fig:nowcasting_results}(b) and (c), we compare four scenarios: (i) without sensitive data, (ii) with sensitive data under no privacy protection, (iii) with sensitive data privatized using the Laplace mechanism, and (iv) with sensitive data privatized using the RR mechanism, along with one baseline (LSTM). 
As shown, \algo{} achieves the best performance when combined with sensitive data under no privacy protection, demonstrating again that sensitive information plays an important role in epidemic modeling. 
When integrated with privatized sensitive data (i.e., with Laplace or RR mechanism), the performance of \algo{} decreases slightly, which is expected. Moreover, compared with the LSTM baseline, the nowcasting projections produced by \algo{} are closer to the ground-truth stable values, especially in the most recent periods. This suggests that our method more accurately captures the underlying temporal trends.

\paragraph*{The Impact of Privacy Budget $\epsilon$}
The Table~\ref{tab:impact_eps_nowcasting} in the Appendix reports the nowcasting performance of our method under different privacy budgets $\epsilon$ with the Laplace mechanism. As shown, our method stably achieves lower error values under different privacy budgets $\epsilon \in \{1, 5, 10\}$, compared to the LSTM baselines ($\epsilon=1$).

\subsection{Reproduction Number ($R_t$) Estimation} 
\begin{figure}
    \centering
    \includegraphics[width=0.5\linewidth]{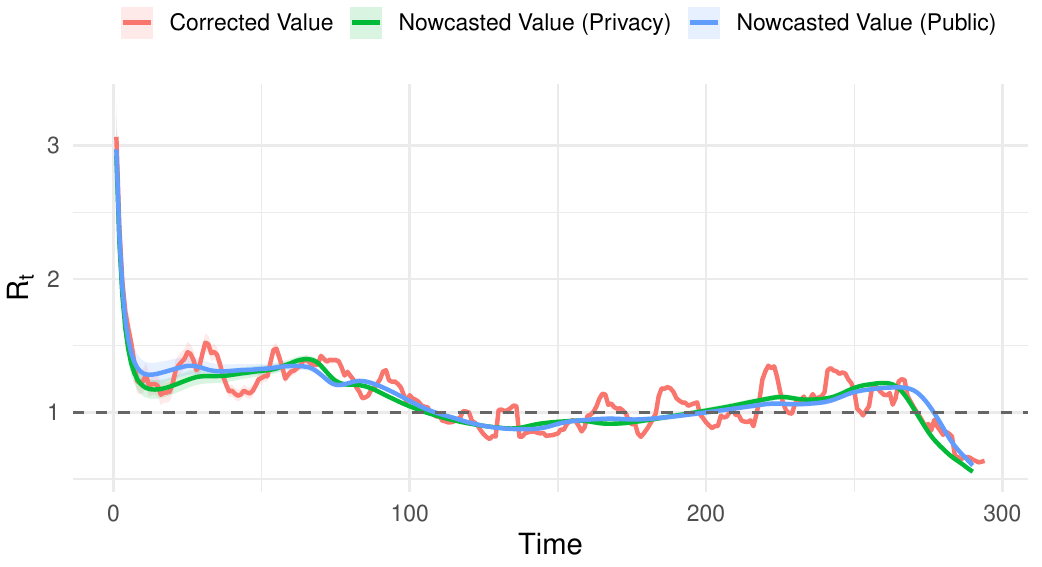}
    \caption{Time-varying $R_t$ estimation by \textit{EpiEstim} with three inputs: (1) Corrected (stable) daily report of new cases; (2) nowcasted daily new cases by the \algo{} under public setting; and (3) nowcasted daily new cases by the \algo{} under private setting.}
    \label{fig:rt}
\end{figure}

Estimation of the effective reproductive number, $R_t$, is important for detecting changes in disease transmission over time. Following~\cite{quevedo2024unveiling}, we estimate the time-varying $R_t$ values by using the epidemiological R package \textit{EpiEstim}~\cite{Cori2013}, assuming an incubation period of 5  days and a serial interval of 6.48  days with a standard deviation equal to 3.83 days~\cite{quevedo2024unveiling}. 
We compare the $R_t$ values computed with three inputs (Figure~\ref{fig:rt}): (1) raw daily report of new cases; (2) nowcasted daily new cases by the \algo{} under public setting (i.e., sensitive data used without any privacy protections); and (3) nowcasted daily new cases by the \algo{} under private setting. 
We observe that the $R_t$ estimates using \algo{} closely follow the estimates from the true case data.


\subsection{Disease transmission model}
A key aspect of \algo{} is that an epidemic transmission model, denoted by $\Minf(\mathbf{p})$, is learned jointly with forecasting or nowcasting.
In principle, different kinds of epidemic models, such as a single patch or multipatch (e.g., age-structured or county-level) SEIRM model, or agent-based model can be used, e.g., \cite{marathe:cacm13, chopra2023differentiable};
the performance can vary depending on the type of epidemic model.
Since datasets are limited, we explore a single patch and an age-structured SEIRM metapopulation model.
As mentioned earlier, for both forecasting and nowcasting (Figures \ref{fig:forecasting_results} and \ref{fig:nowcasting_results}, respectively), the integration of the metapopulation model $\Minf(\mathbf{p})$ within \algo{} gives better performance than the LSTM baseline.
We find, however, that in some cases using the single patch model gives better performance in forecasting than the age-structured model (Table~\ref{tab:calibration} in the Appendix). 
We evaluate how well the learned model $\Minf(\mathbf{p})$ (which is private,  and can be used for other analyses) fits the observational data (Table~\ref{tab:calibration} in the Appendix). 
We find that the age-stratified model achieves lower RMSE in 8 out of 9 comparisons, with relative improvements ranging from 16.80\% to 46.50\%.




\subsection{Potential social distancing interventions}
\label{sec:intervention}



We use a simple SEIRM epidemic model with parameters $p(T')$ learned till time $T'$ (in days), and assume that an intervention, which changes $\beta(t)=(1+x)\beta(T')$, is implemented at time $T'+7\Delta$.
We assume the intervention remains in position starting at $T'+7\Delta$, so $\beta(t)$ is set to
\[
\beta(t) = 
\begin{cases}
    \beta(T'), & t\in \{T'+1, T'+7\Delta-1\}\\
    (1-x)\beta(T'), & t\geq T'+7\Delta
\end{cases}
\]

The goal is to estimate the projected number of infections after time $T'+7\Delta$.
We consider $\Delta=1$ and $2$ weeks, and different values for $x$.
We consider a scenario in which a social distancing type intervention is implemented, and consider $x=-1\%, -5\%$; here we discuss results using the epidemic model learned from \algo{}.
Figure~\ref{fig:counter-factual} shows the number of infections for the learned model (in which $\beta(t)$ keeps changing over the entire learning period), and the four interventions, in which $\beta(t)$ is changed as described above.
As expected, \ul{the number of infections increases with $\Delta$, and decreases with $x$, but the sensitivity with $x$ is much higher.}

\subsection{Uncertainty Quantification} \label{sec:quantile}

Instead of providing only point predictions, \algo{} can be easily modified to incorporate quantile predictions with confidence intervals, which are practically valuable for policymakers. To quantify uncertainty, we use multi-dimensional quantile outputs with different quantile losses, rather than the conventional MSE loss with single-dimensional outputs. Quantile loss is defined as:
\begin{gather*}
     P_{\alpha} = \sum_{i=t}^{T} \alpha\% \max(y_{i} - \hat{y}_{i},0) + (1-\alpha\%) \max(\hat{y}_{i} - y_{i},0), \:\:\nonumber\\ \alpha = 20, 50, 80
\end{gather*}
Here $y_i$ is the ground truth of the target sequence at future $i$-th step, and $\hat{y}_i$ is the corresponding model estimation. Intuitively, $P_{90}$  gives higher penalties on under-estimation, while $P_{10}$ gives higher penalties on over-estimation. Specifically, $\alpha = 50$ represents the median predictions; $\alpha = 20$ gives the 20\% quantile, which gives the lower bound of the confidence interval; and $\alpha = 80$ gives the 80\% quantile, which gives the upper bound of the confidence interval. Figure~\ref{fig:online-setting-quantile} shows quantile predictions with confidence intervals.

\section{Discussion}

We present a powerful framework, \algo{}, for simultaneously predicting epidemic metrics and learning an epidemic model, while integrating diverse and heterogeneous datasets, including sensitive features or datasets that need to be used with differential privacy guarantees.
We show results using \algo{} for two Colombian cities and the USA, and find that adding the financial dataset leads to better performance; for one of the cities, this holds even with DP.
The privately learned epidemic model can be used for any public health task. 

One limitation of the study is the reliance on priors such as the contact matrix $C$ and the age-stratified population vector. We have not conducted a systematic analysis of how inaccuracies in these priors—such as misleading or false information—might influence the effectiveness of \algo{}. In this work, we use data from a well-established prior project~\cite{patchflow}, which has been shown to be informative in previous studies~\cite{venkatramanan2017spatio}. High-quality contact matrices and population vectors enhance the expressive power of meta-population transmission models. However, we acknowledge that erroneous or poorly measured priors may mislead the model during training and could potentially degrade performance.

\algo{} can be leveraged and extended in many directions. One avenue for improvement lies in advancing privacy techniques and expanding the framework to additional privacy models. At present, \algo{} incorporates a plug-in input-perturbation module to privatize the dataset; while this module offers strong portability, it can introduce substantial noise. Developing more sophisticated privacy algorithms~\cite{zhang2022differentially} that achieve a better utility–privacy trade-off without compromising portability would be a promising direction. Second, \algo{} can accommodate heterogeneous data sources, but not all features contribute equally to predictive performance. Incorporating feature-selection or feature-reweighting~\cite{zhang2020doubleensemble} mechanisms may improve downstream accuracy and robustness. Third, although we demonstrated the use of a metapopulation model within \algo{}, the framework can naturally extend to a wide range of epidemic models, including agent-based simulations. Fourth, \algo{} can be applied across a diverse set of application domains beyond those explored in this paper. We anticipate that additional modules or design variations will allow the pipeline to address an even broader set of real-world public health challenges.

To summarize, balancing interpretability, effectiveness, and privacy in public health modelling presents a difficult yet essential challenge. \algo{} takes an important initial step towards demonstrating that these goals can be partially achieved within a unified framework. Looking to the future, designing powerful and responsible public health models will require more calibrated architectures, tailored algorithms, and domain-specific innovations. We expect continued progress and new approaches to emerge in this space.

\section*{Acknowledgments}
{
This research is partially supported by NSF grants CCF-1918656 and CNS-2317193, and DTRA award HDTRA1-24-R-0028, Cooperative Agreement number 6NU50CK000555-03-01 from the Centers for Disease Control and Prevention (CDC) and DCLS, Network Models of Food Systems and their Application to Invasive Species Spread, grant no. 2019-67021-29933 from the USDA National Institute of Food and Agriculture, Agricultural AI for Transforming Workforce and Decision Support (AgAID) grant no. 2021-67021-35344 from the USDA National Institute of Food and Agriculture. The work of R. Tandon is supported in part by NIH Award R01-CA26145701A1, by the US Department of Energy, Office of Science, Office of Advanced Scientific Computing under Award Number DE-SC-ERKJ422, and US NSF under Grants CCF-2100013, CNS-2209951, CNS-2317192. The work of Z. Zhao and B. A. Prakash was partly supported by the NSF
(Expeditions CCF-1918770, CAREER IIS-2028586, Medium IIS-1955883,
Medium IIS-2403240, Medium IIS-2106961), NIH
(1R01HL184139), CDC MInD program, Meta, and Dolby faculty gifts.}

\clearpage

\appendix

\section{Related work}
\label{sec:related}

Even before the pandemic, the value of non-traditional data sources (compared to standard line lists and other epidemiological data) for epidemic forecasting and projection was well recognized; such datasets include different kinds of social media, e.g,~\cite{culotta2010towards, abouzahra2021twitter, chen2014flu},
online search and website logs, e.g.,~\cite{ginsberg2009detecting, mciver2014wikipedia, polgreen2007use}, 
symptomatic online surveys, e.g.,~\cite{smolinski2015flu, salomon2021us},
medical and wellness devices, e.g.,~\cite{miller2018smartphone, leuba2020tracking},
retail and commerce data~\cite{nsoesie2014guess, miliou2021predicting}, detailed mobility traces~\cite{pepe2020covid, klise2021analysis}, detailed disease surveillance, behavioral surveys, and
financial data~\cite{andersen2022consumer, sheridan2020social, alexander2023stay}; see survey~\cite{rodriguez2022data}.
While financial datasets have been used in public health, e.g.,~\cite{andersen2022consumer, sheridan2020social, alexander2023stay}, these have been primarily focused on qualitative trends or understanding changes in specific behaviors, e.g., the level of movements and potential contacts, and less on learning epidemic models.

One area of public health where privacy was considered systematically was in the contact tracing apps developed during the pandemic, e.g.,~\cite{chan:chb21, tran2021health}.
There has been very limited work on public health analyses with differential privacy~\cite{li2024computing, li2022differentially}; however, the broader area of problems like forecasting and learning epidemic models with DP remains open.

\section{Background and Notations}
\label{sec:supplementary}

We briefly describe the background and notation used in the paper.

Many of the datasets we use (and others commonly used in epidemiology) are spatial, and here we focus on zip codes (also referred to as postal codes in some South American cities we consider).
Let $Z$ denote the set of zipcodes.
For a given time stamp $t$ (which would be at the level of days),
let $y_{i,t}$, $m_{i, t}$, $\xsb_{i,t}$ and $\xpriv_{i, t}$ denote the number of infections, number of deaths, tensor of public features, and a private dataset, respectively, for $i \in Z$; we use $\mathbf{y}_{t}$, $\mathbf{m}_{t}$, $\xsbbf_t$ and $\xprivbf_t$ to denote the corresponding tensors (over all groups). Let $\mathbf{y}_{t}=\sum_{i\in Z} y_{i, t}$ and $\mathbf{m}_{t} = \sum_i m_{i, t}$ denote the total number of infections and deaths, respectively, at the city level.


\paragraph*{Forecasting}
The goal in the forecasting problem is to use data till time $T$, namely $\{\mathbf{y}_t\}_{t=1}^T$, $\{\mathbf{m}_t\}_{t=1}^T$, $\{\xsbbf_t\}_{t=1}^T$ and $\{\xprivbf_t\}_{t=1}^T $,
and predict $\{\mathbf{y}_t\}_{T+1}^{T+H}$ and $\{\mathbf{m}_t\}_{T+1}^{T+H}$ i.e.,
the number of infections and deaths during the time period $[T+1, T+H]$, 
where $H$ is the forecast horizon.

\paragraph*{Nowcasting}
Due to reporting errors and delays that can be caused by various factors such as testing artifacts, the preliminarily reported data are always incomplete and subject to retrospective upward corrections~\cite{kamarthi2021back2future}. Therefore, unlike the standard forecasting setting, which assumes that observed values remain stable, we consider a more realistic nowcasting scenario in which the statistics released at each timestamp may undergo continuous revisions until reaching their final, stable versions.

To incorporate these revisions into our formulation, we introduce an additional superscript to indicate the revision index. Let $\{\mathbf{y}^{(t')}_t\}_{t=1}^T$ denote the time series from time 1 to $T$, observed at time $t'$, where $\mathbf{y}^{(t')}_t$ represents the signal at time $t$ as observed at $t'$, and $t' \geq t$.

Focusing on the signal for a given time \(t'\), we define a revision sequence as
\[
\text{seq}(t') = \{\mathbf{y}^{(t')}_{t'}, \mathbf{y}^{(t'+1)}_{t'}, \ldots, \mathbf{y}^{(t_f)}_{t'}\},
\]

where $t_f$ is the final time in our revision dataset. The stability time of the sequence \(\text{seq}(t')\) is the minimal revision time \(r^*\) such that the error between each revised value at later time \(r\) and the final value \(\mathbf{y}^{(t_f)}_{t'}\) is bounded by some \(\epsilon\), i.e.,
\[
\frac{|\mathbf{y}^{(r)}_{t'} - \mathbf{y}^{(t_f)}_{t'}|}{|\mathbf{y}^{(t_f)}_{t'}|} \leq \epsilon \quad \text{for all } r \geq r^*.
\]

To sum up, the goal of the nowcasting is that, at time $T$ (with $T \ll t_f$), given the historical real-time sequences \(\{\{\mathbf{y}^{(t')}_{t''}\}_{t''=1}^{t'}\}_{t'=1}^{T}\), some auxiliary multi-variate sequences, i.e., \(\{\textbf{x}^{pub}_{t'}\}_{t'=1}^{T}\) and \(\{\textbf{x}^{priv}_{t'}\}_{t'=1}^{T}\), and a time window size $w$, revising the recent signals $\textbf{y}^{(T)}_{T},...,\textbf{y}^{(T)}_{T-w}$ to approximate their stable values $\textbf{y}^{(t_f)}_{T},...,\textbf{y}^{(t_f)}_{T-w}$.

\paragraph*{$R_t$ Estimation} For $R_t$ estimation, the task is to estimate time-varying effective reproduction number $\{R_t\}_{t=1}^{T}$ based on a time series of observed incident counts $\{\mathbf{y}_t\}_{t=1}^T$. In this paper, we approach this by using the off-the-shelf epidemiological R package \textit{EpiEstim} following~\cite{quevedo2024unveiling}. The \textit{EpiEstim} models the number of new infections at time as a Poisson (or NB) process, whose mean depends on: (1) past infections, (2) the serial interval distribution, and (3) the current $R_t$. Following~\cite{quevedo2024unveiling}, we adopt an incubation period of 5  days and a serial interval of 6.48  days with a standard deviation equal to 3.83 days. Our task here is to evaluate whether the estimated $R_t$ values change much when using raw infection counts $\{\textbf{y}^{(t_f)}_{t}\}_{t=1}^{T}$ and the nowcasted infection counts ($\{\hat{\textbf{{y}}}^{(T)}_{t}\}_{t=1}^{T}$).

\paragraph*{Epidemic models.}
We consider different kinds of epidemic models, denoted by $\Minf(\mathbf{p})$ (with $\mathbf{p}$ denoting the vector of parameters), including agent-based models, e.g.,~\cite{marathe:cacm13, kerr2021covasim}, and metapopulation models, e.g.,~\cite{mistry2021inferring, venkatramanan2019optimizing}.
A structured metapopulation model represents the transmission of a disease within a set $\mathcal{G}$ of groups; depending on the questions of interest and data, different kinds of groups can be considered, e.g., based on age groups or population in a zip code. 
We describe the simplest such SEIRM metapopulation model, $\Minf(\mathbf{p})$, for the evolution of infections $\mathbf{y}_{t}$ and deaths $\mathbf{m}_{t}$ in Section~\ref{sec:epidemodelgrad}, along with how it is used in our approach.

This can be easily extended to include other states and interventions; we refer to~\cite{marathe:cacm13, kerr2021covasim} for discussion on ABMs.

\section{Methods}

Our \algo{} pipeline consists of three modules described in \S~\ref{sec:algo}: (1) a \parameterNN{} that calibrates the epidemic parameters for the Metapopulation model; (2) a Metapopulation model for epidemic simulation and prediction; and (3) an error correction adapter that further refines the prediction. Additionally, the sensitive dataset is privatized by an input perturbation module described in \S~\ref{sec:privacy}. Because the specific setups for the three datasets considered differ slightly, we describe the method using the Bogota dataset as an illustrative example, while the setups for the other two datasets can be obtained in a straightforward manner following the same pipeline.

\subsection{\algo{} for learning epidemic models and forecasting dynamics}
\label{sec:algo}

\subsubsection{Differentiable meta-population model overview} 
Real-world epidemiological forecasting problems typically require both forecasting accuracy and interpretability~\cite{mathis2024evaluation}. However, existing methods often rely exclusively on either mechanistic models, such as meta-population or agent-based models, or deep neural networks, such as recurrent networks and transformers, making it difficult to achieve both goals within a single model. On the one hand, while mechanistic models offer strong interpretability, they suffer from limited forecasting accuracy due to their inherent model capacity constraints. On the other hand, although deep neural models can provide accurate predictions, their black-box nature offers little interpretability, which is crucial for epidemiologists and decision-makers. 

Several approaches have been proposed that integrate agent-based models with neural networks for parameter calibration~\cite{chopra2022differentiable, anand2024h2abm}. However, these methods often face scalability challenges and are typically limited to micro-level scenarios, such as simulating infections within a hospital. In contrast, epidemiological forecasting also demands models capable of operating at a macro level while maintaining both accuracy and interpretability. To this end, we propose \algo{}, a differentiable meta-population framework that addresses both concerns while ensuring practical applicability in macro-level disease forecasting.

The proposed \algo{} framework consists of three key components:  First, a \textbf{\parameterNN{}}, which takes both public and private data from diverse sources as input and outputs time-varying parameters for the mechanistic model. By leveraging multiple data sources rather than relying solely on the target time series, ParameterNN enables more accurate parameter calibration and better trend extraction.  Second, a \textbf{meta-population model} that captures interactions between different groups, such as age groups or geographic regions. Unlike agent-based models, which operate at a finer scale, the meta-population model offers flexibility in selecting groups at different levels of granularity, enhancing its adaptability for broader forecasting applications.  Third, a \textbf{forecasting adaptor} is designed to correct forecasting predictions adaptively. While the meta-population model, combined with ParameterNN, can lead to over-calibration due to the data richness—resulting in degraded forecasting accuracy—the adaptor mitigates such overfitting issues, improving overall predictive performance.

\subsubsection{Epidemic Model}
\label{sec:epidemodelgrad}
Here, we describe the modeling of the meta-population model. 
A meta-population model is a multi-patch extension of the simple SEIRM model, allowing a more fine-grained simulation of interactions between a set $\mathcal{G}$ of groups, such as the pandemic transmission, in the given region. 
An example is groups correspond to the population in specific age groups; age-stratified data is publicly available~\cite{patchflow} and has been shown to be a reliable signal for epidemic forecasting~\citep{goeyvaerts2010estimating}.

A meta-population model requires two key inputs: a contact matrix $\mathcal{C} \in \mathbb{R}^{m \times m}$ and a population vector $N \in \mathbb{R}^{m}$, where $m=|\mathcal{G}|$ is the number of groups. 
Each entry $C_{i,j} \in \mathcal{C}$ represents the contact probability between the populations in groups $i$ and $j$, while each entry $N^j \in N$ records the population of group $j$. Following~\cite{venkatramanan2019optimizing}, the effective number of infected individuals and the effective population of group $j$ after daily movement are defined as:
\begin{equation}
    I^{\textsc{eff}}_j = \sum_{i} \mathcal{C}_{i,j} I_i, \quad N^{\textsc{eff}}_j = \sum_i \mathcal{C}_{i,j} N_i
\end{equation}
where "\textsc{eff}" (effective) refers to infections accounted including infected individuals between different subpopulations due to transmissibility and mobility. The conditional force of infection for group $j$ is then given by:
\begin{equation}
    \beta^{\textsc{eff}}_j = \beta \frac{I^{\textsc{eff}}_j}{N^{\textsc{eff}}_j}.
\end{equation}
For each group $i \in [1,...,m]$, the epidemic evolution $\mathcal{Z}_i = [S_i, E_i, \allowbreak I_i, R_i, M_i]$ from day $t$ to day $t+1$ is formulated as:
\begin{equation}
\begin{aligned}
    \mathcal{Z}_i(t+1) &= \mathcal{Z}_i(t) + \Delta {\mathcal{Z}_i(t)}, \quad where \\
    \Delta {S_i(t)} &= -\sum_{j=1}^{m} C_{i,j} \beta^{\textsc{eff}}_j S_i(t) + \delta(t) R_i(t),\\
    \Delta {E_i(t)} &= \sum_{j=1}^{m} C_{i,j} \beta^{\textsc{eff}}_j S_i(t) - \alpha(t) E_i,\\
    \Delta {I_i(t)} &= \alpha(t) E_i(t) - (\gamma(t) + \eta(t)) I_i(t), \\
    \Delta {R_i(t)} &= \gamma(t) I_i(t) + (1-\delta(t)) R_i(t), \\
    \Delta {M_i(t)} &= \eta(t) I_i(t). 
    \label{eqn:transmission}
\end{aligned}
\end{equation}
where our forecasting objective $\Delta I_i(t)$ represents the number of newly infected individuals in age group $i$ at time $t$. Running the meta-population model for $T$ time steps, it outputs the aggregated new cases over all the age groups for the entire training period:
\begin{equation}
    \hat{y}_t := \sum_{i=1}^m \Delta I_i(t), \forall 1 \leq t \leq T.
\end{equation}

\subsubsection{\parameterNN}
The meta-population model incorporates a large number of time-varying hyperparameters, such as $\alpha(t)$, $\gamma(t)$, and $\eta(t)$ for the pandemic transmission simulation. Despite the challenge brought by these complex parameters, accurate calibration time-dependently becomes even more challenging, especially when the total time step $T$ is large. Consequently, the vast search space makes calibrating these epidemic parameters highly challenging, often requiring manual efforts or relying on traditional calibration strategies such as EAKF~\citep{cao2022eakf}, which frequently fail. Motivated by the recent success of differentiable calibration methods~\citep{chopra2022differentiable}, we propose a DNN-based approach that predicts epidemic parameters from given datasets. Intuitively, our goal is to train a neural network model, $f_{\theta_{epi}}$, capable of capturing pandemic trends through these predicted epidemic parameters.

In our setup, we incorporate both the public datasets, such as city-level Google Health Trends, Google Mobility Dataset, and Infection Statistics, and the private ones are more fine-grained zip-level datasets from credit transaction data. To simultaneously handle both granularity and generate time-varying epidemic parameters, our neural network $f_{\theta_{epi}}$ is designed to be a two-encoder-based sequence model so that it can encode the zip-level information and the City-level information separately. Let $\mathcal{D}_{tran}$ denote the private transaction dataset, and $\mathcal{D}_{public}$ denote the merged public dataset, we get the predicted epidemic parameters $p(t) := \{\alpha(t), \gamma(t),\eta(t), ...\}$ as follows:
\begin{equation}
    \{p(t)\}_{t=1}^{T} = f_{\theta_{epi}} (\mathcal{D}_{tran}, \mathcal{D}_{public}).
\end{equation}

To improve the learning stability in the training process, we only update $p(t)$ on a weekly basis, i.e., $p(t) = p(t')$ for $t, t'\in\{7i+1,\ldots,7(i+1)\}$, for all $i$, though the forecasts and epidemic model runs at a daily level. More details on the model architecture are provided in the Appendix.

\subsubsection{Error Correction Adapter}
While the proposed \parameterNN{} enhances interpretability for historical calibration, it tends to suffer from reduced forecasting accuracy due to the lack of calibration in the horizon time series.
To address this issue, we design a novel correction adapter module that locally aligns the simulation with the ground-truth epidemic curves to refine the simulation in the horizon window. Formally, given a simulation sequence $\hat{y}$ of length $T$, the error correction adapter first partitions the sequence into overlapping contiguous chunks of size $k$. Each chunk, defined as $\text{chunk}_j = \{\hat{y}_j, \hat{y}_{j+1}, ..., \hat{y}_{j+k-1}\}$, captures a short time window of the simulation for localized correction. This approach allows the adapter to focus on refining small segments of the sequence rather than attempting a global adjustment all at once. Next, the adapter predicts a corrected value for each chunk based on its historical simulation data, using a network $g_{\theta}$ parametrized by $\theta$, i.e., $\tilde{y}_j = g_{\theta}(\text{chunk}_j)$. By applying this correction iteratively across all chunks, we obtain a refined sequence of adjusted values, $\{\tilde{y}_j\}_{j=1}^{T-k}$, which better reflects the expected epidemic trajectory.

\subsubsection{Loss Calculation}

Given the simulation $\{\hat{y}_t\}_{t=1}^{T}$ and the refined simulation $\{\tilde{y}_t\}_{t=1}^{T-k}$, we can calculate two loss functions:

\begin{equation}
    \mathcal{L}_{MetaPop} =\frac{1}{T} \sum_{t=1}^{T} \|\hat{y}_t - y_t\|_2,
    \label{eqn:gradmeta_loss}
\end{equation}
and
\begin{equation}
    \mathcal{L}_{Adapter} = \frac{1}{T-k} \sum_{t=1}^{T-k} \|\tilde{y}_t - y_{t+k}\|_2,
    \label{eqn:adapter_loss}
\end{equation}
respectively. Combining the two losses together with a linear rate $\alpha$, we get
\begin{equation}
    \mathcal{L} = \alpha\cdot\mathcal{L}_{MetaPop} + (1-\alpha) \cdot \mathcal{L}_{Adapter}.
    \label{eqn:combined_loss}
\end{equation}

To train the whole pipeline, one can simply back-propagate with the gradients of loss value defined in Equation~\ref{eqn:combined_loss}.

However, it is generally challenging to select an optimal $\alpha$ value. Moreover, using a constant $\alpha$ is more likely to result in convergence issues. Therefore, instead of maintaining a constant $\alpha$ throughout all training epochs, we propose defining $\alpha_t$ as a function of the training epoch $t$. Specifically,
\begin{equation}
    \alpha_t = \frac{t}{T},
\end{equation}
where $T$ denotes the total training epochs. Intuitively, $\alpha_t$ is a linearly increasing function of the training epoch $t$. 
During the early stages of training, the weight assigned to the \parameterNN{} module is relatively larger. Gradually, this weight shifts toward the adapter module. 
This design is based on our empirical observation that the \parameterNN{} module converges more slowly than the adapter module. 
Assigning a larger weight to this module during the early training stages promotes more stable loss reduction.

\subsection{Privacy mechanisms}
\label{sec:privacy}

Due to the mixed nature of our public and private datasets, we employ two input perturbation techniques to ensure privacy. 
Our private transaction data, including zip codes and transaction amounts, are both considered sensitive and require privacy protection. 
This means the location (zip code) and the amount of each individual transaction must be kept private. 
Furthermore, because each merchant may have transactions at multiple locations, a composition mechanism is necessary to provide merchant-level privacy. 
As shown in our pipeline figures, input perturbation is applied to the transaction data, which is then used for training.

\subsubsection{Privacy via the Laplace mechanism} 

We use transaction data aggregated by amount (for specific categories) at the level of zip codes and weeks.
This aggregated dataset, denoted by table $\Dtran$, consists  of $M$ rows, where row $i$ represents a postal code $P_i$ (for a total of $M$ postal codes: $P_1, \ldots, P_M$). 
$\Dtran$ has $209$ columns, each representing a specific week in our time series of historical transaction data. The value in cell $\Dtran_{ij}$ represents the total amounts of transactions in postal code $P_i$ during week $j$. We clip each original transaction amount by a pre-determined maximum amount of $C$, such that any transaction amount that is greater than $C$ will be treated as $C$.

To calculate the sensitivity of table $\Dtran$, note that removing one merchant from the mock data can remove up to $K = 627$ records in the worst case, which makes the global $\ell_1$-sensitivity of $\Dtran$ to be $627 \times C$. 
Therefore, adding a Laplacian noise with magnitude $\frac{627 \times C}{\epsilon}$ to each cell of $\Dtran$ satisfies $\epsilon$-DP.

\subsubsection{Privacy via label DP using Randomized Response}


We consider another approach for input perturbation based on Label DP~\cite{ghazi2021deep}, which aims to provide a differential privacy guarantee only for the specific label. 
Informally, the goal of label DP is to ensure the influence of any individual data entry's label on the trained model is limited, in this case, the specific pair of merchant ID and postal code. 
This limits the ability of an adversary to infer the true label of any given data point, thereby protecting sensitive information while maintaining the utility of the overall model.
Formally, the notion of Label DP is defined as:

\begin{dfn}[$\{\epsilon, \delta \}$ Label Differential Privacy]
    For all two neighboring datasets $\mathcal{D},\mathcal{D}'$ that differs in the label of a single data entry, and for any subset $\mathcal{S}$ of outputs of randomized algorithm $\mathcal{M}$, the algorithm $\mathcal{M}$ is $(\epsilon,\delta)$-Label Differentially Private (Label DP) for $\epsilon\ge 0$ and $\delta\in[0,1]$ if:
    \begin{align*}
    \mathbb{P}[\mathcal{M}(\mathcal{D})\in \mathcal{S}]\le e^{\epsilon}\cdot\mathbb{P}[\mathcal{M}(\mathcal{D}')\in \mathcal{S}]+\delta.
   \end{align*}
\end{dfn} 


The idea here is to privatize the zip codes and the transaction data separately to provide event-level privacy and use composition to provide merchant-level privacy. 
First, we implement randomized responses on the zip codes with a privacy budget of $\epsilon_{\text{zip}}$. 
Next, to preserve the privacy of the transaction data (transaction details), we used both quantization and randomized response. 
Based on the bin size $B$, we divided the entire transaction data into $B$ bins and implemented a randomized response with a privacy budget of $\epsilon_{\text{trans}}$. 
This proposed methodology provides $\epsilon_{\text{event}}$ event-level privacy with basic composition as:
\begin{align}
    \epsilon_{\text{event}} = \epsilon_{\text{zip}}+\epsilon_{\text{trans}}
\end{align}




\paragraph*{Randomized response with quantization.}
We essentially propose two different privacy budgets for zip code and transaction details. 
This process involves reporting the zip code with a probability of $\frac{e^{\epsilon_{\text{zip}}}}{e^{\epsilon_{\text{zip}}} + k_1 -1 }$ and reporting another random zip code uniformly with a probability of $\frac{1}{e^{\epsilon_{\text{zip}}} + k_1 -1 }$. 
Here, $k_1$ represents the cardinality of the set of all possible zip codes. 
For transaction details, we first quantize the data into $B$ bins and then apply randomized response with a privacy budget of $\epsilon_{\text{trans}}$. 
Without a doubt, the number of bins $B$ will influence the utility of the data; by setting $B=1$ will preserve perfect privacy on transaction details. 
For practical reasons, $B$ is pre-determined to be 50, although this parameter can be tuned for different performance requirements or datasets. 
Therefore, this mechanism returns the true quantized transaction data with probability $\frac{e^{\epsilon_{\text{trans}}}}{e^{\epsilon_{\text{zip}}} + B -1 }$ and reports another random quantized transaction data uniformly with probability of $\frac{1}{e^{\epsilon_{\text{trans}}} + B -1 }$.
However, this method does not rely on prior knowledge of the dataset, specifically does not rely on the distribution of zip codes and transaction details. 

\smallskip



\paragraph*{Merchant-level Privacy.} 
For this notion of privacy, we assume $D\sim D'$ if they differ in entries corresponding to exactly one merchant.
To achieve merchant-level privacy, we have to take into account the composition cost due to the multiple appearances of each unique merchant in the transaction dataset. Since there are $209$ weeks and each merchant has at most $3$ postal codes, a merchant may contain up to $K=209 \times 3 = 627$ transaction entries. 

\medskip
\noindent 




\paragraph*{Reduction in privacy cost using sub-sampling.}
Sub-sampling offers an additional avenue for enhancing privacy. 
Basic composition dictates a merchant-level privacy budget of $K\epsilon_{event}$. 
However, by selecting only $\alpha$ fraction of transactions for each zip code and date, the privacy budget is reduced to $K \alpha \epsilon_{event}$ \cite{ghazilabeldp}. 
In order to achieve $\epsilon = 1$, we choose $\epsilon_{event} = 0.007$ for $\alpha = 0.2$ for practicality. 

\subsection{Experimental Setups}
\subsubsection{Heterogeneous Data Sources} 
\label{sec:data}
In the experiments, we consider three locations: Bogot\'a and Medell\'in, and the USA for the pandemics of COVID-19. 

\paragraph*{Bogota \& Medellin.} For the Bogota and the Mdellin, we collected datasets from three sources: (1) Daily cases and deaths data provided by the National Institute of Health of Colombia, (2) Search trends for symptoms from Google Heath Trends, with keywords such as "covid-19 en Colombia", "covid-19 vacuna", "covid-19 hoy", and "covid-19 Bogot\'a", and (3) Google Mobility Dataset with 6 features relevant to mobility changes in different scenarios. 
Apart from these public datasets, we also consider the transaction dataset released by data.org~\cite{pet-challenge}. Note that the transaction data provided for this research is supplied as-is, with no warranties or guarantees, and the data provider holds no liability for its accuracy, completeness, or any consequences arising from its use. Each entry records the zip code, category, time, and spending amount of transactions for the given location. To utilize this transaction dataset, we preprocess it by aggregating the sum of the spending amounts across all categories, resulting in a dataset relevant only to zip code and time. We consider a default training period of 42 weeks, which starts from 2020-04-12. The prediction horizon is set as 28 days. For the testing stage, we evaluate the model's performance by comparing infection predictions (i.e., 28 days following the training stage) with the ground-truth infection values. 

\paragraph*{USA.} For the USA dataset, we follow~\cite{chopra2022differentiable} to construct the input features, which include: (1) symptomatic survey signals collected from Facebook; (2) symptom-related search trends from Google Health Trends; (3) hospitalization counts; and (4) weighted Influenza-Like Illness (wILI) rates collected by the CDC through the Outpatient Influenza-Like Illness Surveillance Network (ILINet). We treat hospitalization information (i.e., the “Hospitalized Increase Count”) released by the U.S. Department of Health \& Human Services as private. 
 
Table~\ref{tab:dataset} summarizes the datasets used in our experiments.

\subsubsection{Metrics} 
For performance evaluation, we use three popular metrics, including root mean square error (RMSE), mean absolute error (MAE), and mean absolute percentage error (MAPE). For any given prediction $\bm{y}'_t$ and the corresponding ground-truth values $\bm{y}_t$ for an interval $[t_1, t_2]$, the RMSE for this interval is formally written as $\sqrt{\frac{1}{t_2-t_1+1}\sum_{t=t_1}^{t_2} (\bm{y}_t-\bm{y}_t')^2}$, the MAE is $\frac{1}{t_2-t_1+1}\sum_{t=t_1}^{t_2} |\bm{y}_t-\bm{y}_t'|$, and the MAPE is $\frac{1}{t_2-t_1+1}\sum_{t=t_1}^{t_2} |\frac{\bm{y}_t-\bm{y}_t'}{\bm{y}_t}|$.

\subsubsection{Implementation details.}
All of our experiments are conducted on a server with 2 $\times$ A6000 GPU. To ensure learning stability, we set a learning rate of $5e^{-4}$ during the three training stages, and to achieve better convergence, we use 2,000 epochs for the first stage, and 500 epochs for the remaining two stages.

\subsubsection{Baseline Method}
\label{sec:lstm}

\begin{figure}[!h]
    \centering
    \includegraphics[width=\linewidth]{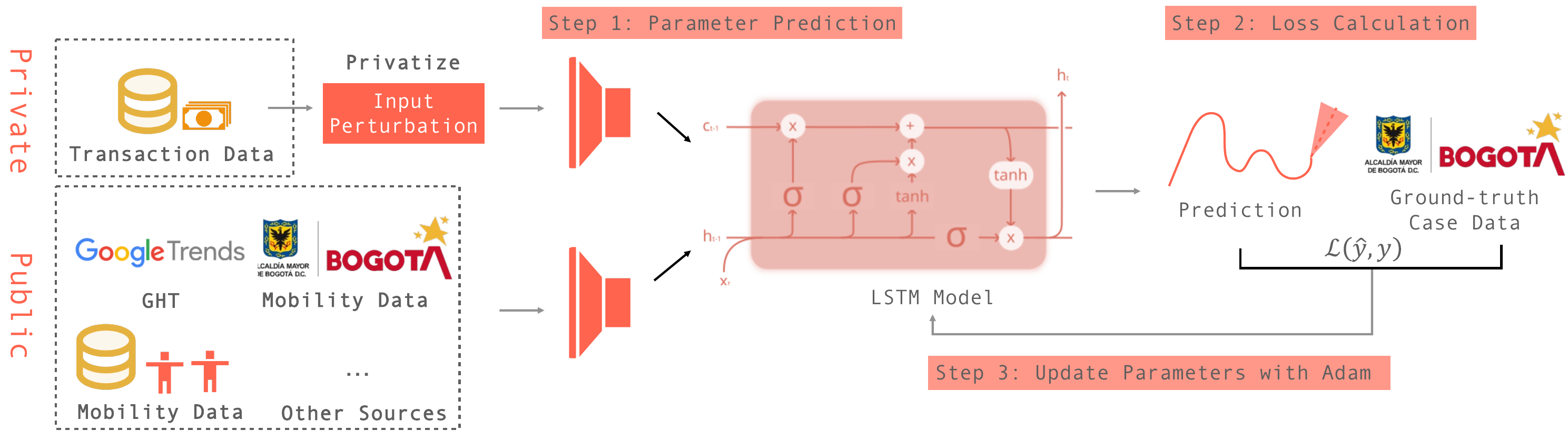}
\caption{Forecasting using the LSTM technique: we use a multi-encoder structure, which allows different types of datasets (whose shapes are not always consistent) to be incorporated. 
Transaction data is incorporated through a separate encoder. Predictions are generated directly by the LSTM in an autoregressive manner.
}
    \label{fig:pipeline_pets_lstm}
\end{figure}

Here, we consider an  LSTM-based method for forecasting (Figure~\ref{fig:pipeline_pets_lstm}).
This is a pure machine learning method, in contrast with our main approach, which couples a neural network with an epidemic model.
We employ a model similar to the LSTM model developed in \cite{adiga2021all} which has been deployed within an ensemble framework and tested extensively in the context of COVID-19 and influenza forecasting. 
The model is simple and consists of one LSTM layer with hidden size 32, one dense layer with hidden size 16, a rectified linear unit activation function, and one dropout layer (dropout rate of 0.2). The output layer is a dense layer with linear activation and $L_2$ kernel regularization with a 0.01 penalty factor. The input dataset was first segmented into several continuous five-timestamp-chunks, where the first four timestamps are used as features and the last timestamp is used as the prediction target. 

We also implement a similar two-encoder-structure for the LSTM, as shown in Figure~\ref{fig:pipeline_pets_lstm}, to handle the two heterogeneous input datasets. 
The first encoder processes the zip-level privatized transaction dataset. 
The second encoder processes public city-level datasets.
The embeddings from both encoders are concatenated and passed to the decoder, which produces the forecasts.

\begin{figure}[!h]
    \centering
    \includegraphics[width=0.8\columnwidth]{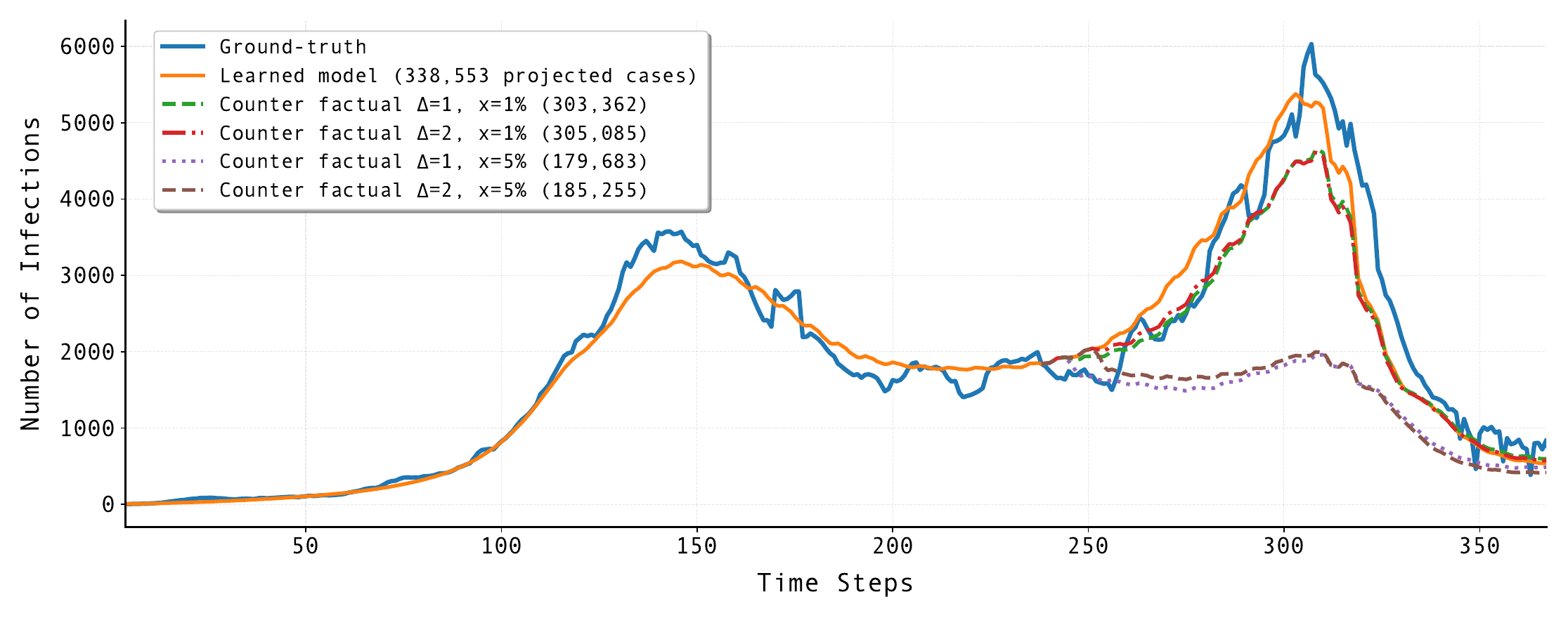}
\caption{
Analysis of interventions: The orange line shows the fit with the trained model using ground-truth data (blue line) for 52 weeks.
The model trained at $T=231$ days is used for evaluating four interventions involving social distancing, which are implemented $\Delta=1,2$ weeks after $T$ and result in a reduction in $\beta$ by $x=1\%, 5\%$.
} 
    \label{fig:counter-factual}
\end{figure}

\begin{figure}[!h]
    \centering
    \includegraphics[width=0.8\linewidth]{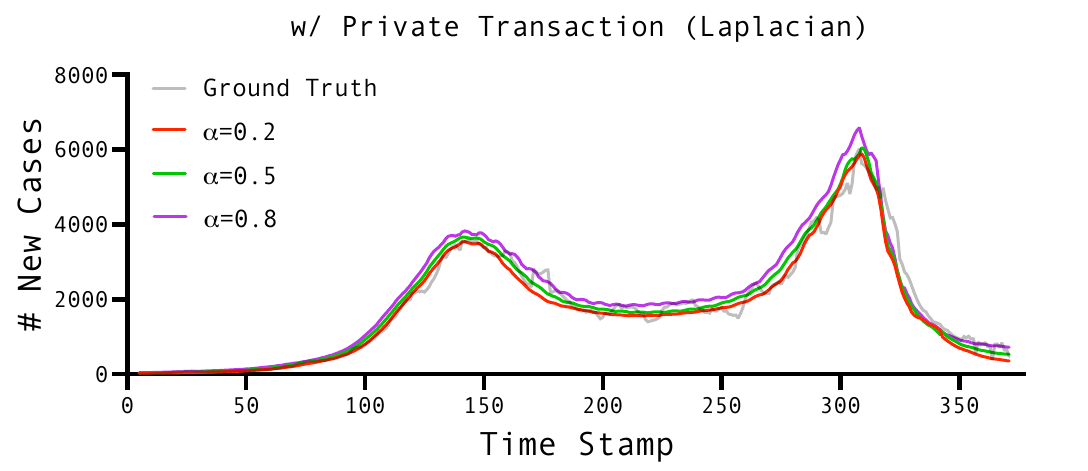}
    \caption{Performance of learning with quantile loss. We use three quantiles $\alpha = 20, 50, 80$.}
    \label{fig:online-setting-quantile}
\end{figure}

\begin{table}[!h]
    \centering
    \caption{Statistical Description of the datasets used in the Experiments.}
    \label{tab:dataset}
    \begin{tabular*}{\textwidth}{c|ccccccc}
    \toprule
        Dataset Name & Data Source & Temporal Granularity  & \# Features & Target\\
    \midrule
         Bogota & Colombia NIH & 343 (days)  & 11 & Infection Counts\\
         Medellin & Colombia NIH & 343 (days) & 11 & Infection Counts\\
         USA & CDC & 602 (days) & 14 & Mortality\\
         \bottomrule
    \end{tabular*}
\end{table}

\begin{table}[!h]
    \centering
    \caption{Impact of $\epsilon$ on the forecasting performance using the Laplacian mechanism on the Bogota dataset.
}
\label{tab:impact_eps}
    \begin{tabular*}{\textwidth}{@{\extracolsep\fill}c|ccccc}
    \toprule
        Metrics & LSTM (Baseline, $\epsilon=1$) & $\epsilon=\infty$ (No Privacy) & $\epsilon = 1$ & $\epsilon = 5$ & $\epsilon = 10$ \\
         \midrule
         RMSE & 1357.79 & 914.27 & 900.68 & 876.52 & 873.27\\
         MAE & 1159.35 & 753.17 & 735.34 & 734.12 & 720.26  \\
         MAPE & 44.69 & 32.20 & 31.43 & 31.27 & 30.33 \\
         \bottomrule
    \end{tabular*}
\end{table}

\begin{table}[!h]
    \centering
    \caption{Impact of $\epsilon$ on the nowcasting performance using the Laplacian mechanism on the Bogota dataset.}
    \label{tab:impact_eps_nowcasting}
    \begin{tabular*}{\textwidth}{@{\extracolsep\fill}c|ccccc}
    \toprule
        Metrics & LSTM (Baseline, $\epsilon=1$) & $\epsilon=\infty$ (No Privac) & $\epsilon = 1$ & $\epsilon = 5$ & $\epsilon = 10$ \\
         \midrule
         RMSE & 2080.97 & 1504.92 & 1655.24 & 1266.49 & 1236.67\\
         MAE & 1757.57 & 1342.66  & 1278.06 & 1251.49 & 1123.56 \\
         MAPE & 57.34 & 29.50 & 34.52 & 33.56 & 31.25 \\
         \bottomrule
    \end{tabular*}
\end{table}

\begin{table}[!h]
    \centering
    \caption{Performance (RMSE) for the calibration and the forecasting.}
    \label{tab:calibration}
    \begin{tabular*}{\textwidth}{@{\extracolsep\fill}c|ccccccc}
    \toprule
        Dataset & Model & \multicolumn{2}{c}{Public} & \multicolumn{2}{c}{Lap} & \multicolumn{2}{c}{RR} \\
        & & Calibration & Forecasting & Calibration & Forecasting & Calibration & Forecasting \\
         \midrule
        \multirow{2}*{Bogota} & \texttt{Age-stratified} & 267.53 & 914.27 & 287.97 & 900.68 & 374.98 & 1075.18 \\
         & \texttt{Single-patch} & 499.99 & 705.74 & 499.69 & 705.62 & 562.28 & 1027.73 \\
        \multirow{2}*{Medellin} & \texttt{Age-stratified} & 108.71 & 523.85 & 140.83 & 776.15 & 86.77 & 649.80 \\
        & \texttt{Single-patch} & 156.29 & 567.89 & 205.67 & 330.47 & 104.28 & 812.29\\
        \multirow{2}*{USA} & \texttt{Age-stratified} & 133.45 & 4683.84 & 156.91 &  895.80  & 162.67 & 1014.11\\
        & \texttt{Single-patch} & 175.25 & 1385.25 & 188.57 & 1342.19 & 142.45	& 1278.69 \\
         \bottomrule
    \end{tabular*}
\end{table}

\bibliographystyle{unsrt}  
\bibliography{references}

@inproceedings{adiga2021all,
  title={All models are useful: Bayesian ensembling for robust high resolution covid-19 forecasting},
  author={Adiga, Aniruddha and Wang, Lijing and Hurt, Benjamin and Peddireddy, Akhil and Porebski, Przemyslaw and Venkatramanan, Srinivasan and Lewis, Bryan Leroy and Marathe, Madhav},
  booktitle={Proceedings of the 27th ACM SIGKDD Conference on Knowledge Discovery \& Data Mining},
  pages={2505--2513},
  year={2021}
}

@article{ghazi2021deep,
  title={Deep learning with label differential privacy},
  author={Ghazi, Badih and Golowich, Noah and Kumar, Ravi and Manurangsi, Pasin and Zhang, Chiyuan},
  journal={Advances in neural information processing systems},
  volume={34},
  pages={27131--27145},
  year={2021}
}

@inproceedings{chopra2023differentiable,
  title={Differentiable Agent-based Epidemiology},
  author={Chopra, Ayush and Rodr{\'\i}guez, Alexander and Subramanian, Jayakumar and Quera-Bofarull, Arnau and Krishnamurthy, Balaji and Prakash, B Aditya and Raskar, Ramesh},
  booktitle={Proceedings of the 2023 International Conference on Autonomous Agents and Multiagent Systems},
  pages={1848--1857},
  year={2023}
}

@misc{patchflow,
author = {Przemyslaw Porebski and Srini Venkatramanan}, 
year = 2021,
title = {PatchFlow v1.0 mobility networks}, 
doi = {10.5281/zenodo.4495680}
}

@article{mistry2021inferring,
  title={Inferring high-resolution human mixing patterns for disease modeling},
  author={Mistry, Dina and Litvinova, Maria and Pastore y Piontti, Ana and Chinazzi, Matteo and Fumanelli, Laura and Gomes, Marcelo FC and Haque, Syed A and Liu, Quan-Hui and Mu, Kunpeng and Xiong, Xinyue and others},
  journal={Nature communications},
  volume={12},
  number={1},
  pages={323},
  year={2021},
  publisher={Nature Publishing Group UK London}
}

@article{rodriguez2022data,
  title={Machine learning for data-centric epidemic forecasting},
  author={Rodr{\'\i}guez, Alexander and Kamarthi, Harshavardhan and Agarwal, Pulak and Ho, Javen and Patel, Mira and Sapre, Suchet and Prakash, B. Aditya},
  journal={Nature Machine Intelligence},
  year={2024},
doi={https://www.nature.com/articles/s42256-024-00895-7}
}

@article{goeyvaerts2010estimating,
  title={Estimating infectious disease parameters from data on social contacts and serological status},
  author={Goeyvaerts, Nele and Hens, Niel and Ogunjimi, Benson and Aerts, Marc and Shkedy, Ziv and Van Damme, Pierre and Beutels, Philippe},
  journal={Journal of the Royal Statistical Society Series C: Applied Statistics},
  volume={59},
  number={2},
  pages={255--277},
  year={2010},
  publisher={Oxford University Press}
}

@article{kamarthi2021back2future,
  title={Back2future: Leveraging backfill dynamics for improving real-time predictions in future},
  author={Kamarthi, Harshavardhan and Rodr{\'\i}guez, Alexander and Prakash, B Aditya},
  journal={arXiv preprint arXiv:2106.04420},
  year={2021}
}

@inproceedings{ghazilabeldp,
  title={LabelDP-Pro: Learning with Label Differential Privacy via Projections},
  author={Ghazi, Badih and Huang, Yangsibo and Kamath, Pritish and Kumar, Ravi and Manurangsi, Pasin and Zhang, Chiyuan},
  booktitle={The Twelfth International Conference on Learning Representations}
}

@inproceedings{10.1145/1065167.1065184,
author = {Blum, Avrim and Dwork, Cynthia and McSherry, Frank and Nissim, Kobbi},
title = {Practical Privacy: The SuLQ Framework},
year = {2005},
isbn = {1595930620},
publisher = {Association for Computing Machinery},
address = {New York, NY, USA},
url = {https://doi.org/10.1145/1065167.1065184},
doi = {10.1145/1065167.1065184},
pages = {128–138},
numpages = {11},
location = {Baltimore, Maryland},
series = {PODS '05}
}

@article{dwork2011differential,
  title={Differential privacy},
  author={Dwork, Cynthia},
  journal={Encyclopedia of Cryptography and Security},
  pages={338--340},
  year={2011},
  publisher={Springer}
}

@article{dwork:fttcs14,
  title={The algorithmic foundations of differential privacy.},
  author={Dwork, Cynthia and Roth, Aaron},
  journal={Foundations and Trends in Theoretical Computer Science},
  volume={9},
  number={3-4},
  pages={211--407},
  year={2014}
}

@inproceedings{nissim:stoc07,
author = {Nissim, Kobbi and Raskhodnikova, Sofya and Smith, Adam},
title = {Smooth Sensitivity and Sampling in Private Data Analysis},
year = {2007},
isbn = {9781595936318},
publisher = {Association for Computing Machinery},
address = {New York, NY, USA},
url = {https://doi.org/10.1145/1250790.1250803},
doi = {10.1145/1250790.1250803},
booktitle = {Proceedings of the Thirty-Ninth Annual ACM Symposium on Theory of Computing},
pages = {75–84},
numpages = {10},
location = {San Diego, California, USA},
series = {STOC '07}
}

@inproceedings{cummings2018role,
  title={The role of differential privacy in gdpr compliance},
  author={Cummings, Rachel and Desai, Deven},
  booktitle={FAT’18: Proceedings of the Conference on Fairness, Accountability, and Transparency},
  volume={20},
  year={2018}
}

@misc{nsf-pet,
author = {NSF-NIST},
title = {PETs Prize Challenge: Phase 1.
Data Track B: Pandemic Forecasting
Transforming Pandemic Response and Forecasting},
year = 2023
}

@misc{harrison:pet-challenge,
author = {Harrison, G and Chen, J and Mortveit, H and Hoops, S and Porebski, P and Xie, D and Wilson, M and Bhattacharya, P and Vullikanti, A and Xiong, L and Marathe, M}, 
year = 2023, 
title = {Synthetic Data To Support US-UK Prize Challenge For Developing Privacy Enhancing Methods: Predicting Individual Infection Risk During A Pandemic}, 
doi = {10.18130/V3/ZOG1FF}
}

@misc{pet-challenge,
title = { Privacy {E}nhancing {T}echnologies ({PETs}) for Public Health Challenge },
note = {\url{https://data.org/initiatives/pets-challenge/}},
year={2024}
}

@article{kerr2021covasim,
  title={{C}ovasim: {A}n {A}gent-{B}ased {M}odel of {COVID}-19 {D}ynamics and {I}nterventions},
  author={Kerr, Cliff C and Stuart, Robyn M and Mistry, Dina and Abeysuriya, Romesh G and Rosenfeld, Katherine and Hart, Gregory R and N{\'u}{\~n}ez, Rafael C and Cohen, Jamie A and Selvaraj, Prashanth and Hagedorn, Brittany and others},
  journal={PLOS Computational Biology},
  volume={17},
  number={7},
  pages={e1009149},
  year={2021},
  publisher={Public Library of Science San Francisco, CA USA}
}

@Article{marathe:cacm13,
  Title                    = {Computational Epidemiology},
  Author                   = {Madhav Marathe and Anil Vullikanti},
  Journal                  = {Communications of the ACM},
  Year                     = {2013},
  volume = {56},
  number = {7},
  pages = {88--96},
}

@article{venkatramanan2019optimizing,
  title={Optimizing spatial allocation of seasonal influenza vaccine under temporal constraints},
  author={Venkatramanan, Srinivasan and Chen, Jiangzhuo and Fadikar, Arindam and Gupta, Sandeep and Higdon, Dave and Lewis, Bryan and Marathe, Madhav and Mortveit, Henning and Vullikanti, Anil},
  journal={PLoS computational biology},
  volume={15},
  number={9},
  pages={e1007111},
  year={2019},
  publisher={Public Library of Science San Francisco, CA USA}
}

@article{cao2022eakf,
  title={EAKF-Based parameter optimization using a hybrid adaptive method},
  author={Cao, Lige and Wu, Xinrong and Han, Guijun and Li, Wei and Wu, Xiaobo and Wu, Haowen and Li, Chaoliang and Li, Yundong and Zhou, Gongfu},
  journal={Monthly Weather Review},
  volume={150},
  number={11},
  pages={3065--3080},
  year={2022}
}

@article{mathis2024evaluation,
  title={Evaluation of FluSight influenza forecasting in the 2021--22 and 2022--23 seasons with a new target laboratory-confirmed influenza hospitalizations},
  author={Mathis, Sarabeth M and Webber, Alexander E and Le{\'o}n, Tom{\'a}s M and Murray, Erin L and Sun, Monica and White, Lauren A and Brooks, Logan C and Green, Alden and Hu, Addison J and Rosenfeld, Roni and others},
  journal={Nature communications},
  volume={15},
  number={1},
  pages={6289},
  year={2024},
  publisher={Nature Publishing Group UK London}
}

@inproceedings{anand2024h2abm,
  title={H2ABM: Heterogeneous Agent-based Model on Hypergraphs to Capture Group Interactions},
  author={Anand, Vivek and Cui, Jiaming and Heavey, Jack and Vullikanti, Anil and Prakash, B Aditya},
  booktitle={Proceedings of the 2024 SIAM International Conference on Data Mining (SDM)},
  pages={280--288},
  year={2024},
  organization={SIAM}
}

@inproceedings{chopra2022differentiable,
  title={Differentiable Agent-based Epidemiology},
  author={Chopra, Ayush and Rodr{\'\i}guez, Alexander and Subramanian, Jayakumar and Krishnamurthy, Balaji and Prakash, B Aditya and Raskar, Ramesh},
  booktitle={AAMAS},
  year={2023}
}

@article{andersen2022consumer,
  title={Consumer responses to the COVID-19 crisis: Evidence from bank account transaction data},
  author={Andersen, Asger Lau and Hansen, Emil Toft and Johannesen, Niels and Sheridan, Adam},
  journal={The Scandinavian Journal of Economics},
  volume={124},
  number={4},
  pages={905--929},
  year={2022},
  publisher={Wiley Online Library}
}

@article{sheridan2020social,
  title={Social distancing laws cause only small losses of economic activity during the COVID-19 pandemic in Scandinavia},
  author={Sheridan, Adam and Andersen, Asger Lau and Hansen, Emil Toft and Johannesen, Niels},
  journal={Proceedings of the National Academy of Sciences},
  volume={117},
  number={34},
  pages={20468--20473},
  year={2020},
  publisher={National Acad Sciences}
}

@article{alexander2023stay,
  title={Do stay-at-home orders cause people to stay at home? Effects of stay-at-home orders on consumer behavior},
  author={Alexander, Diane and Karger, Ezra},
  journal={Review of Economics and Statistics},
  volume={105},
  number={4},
  pages={1017--1027},
  year={2023},
  publisher={MIT Press One Rogers Street, Cambridge, MA 02142-1209, USA journals-info~…}
}

@inproceedings{culotta2010towards,
  title={Towards detecting influenza epidemics by analyzing Twitter messages},
  author={Culotta, Aron},
  booktitle={Proceedings of the first workshop on social media analytics},
  pages={115--122},
  year={2010}
}

@article{abouzahra2021twitter,
  title={Twitter vs. Zika—The role of social media in epidemic outbreaks surveillance},
  author={Abouzahra, Mohamed and Tan, Joseph},
  journal={Health Policy and Technology},
  volume={10},
  number={1},
  pages={174--181},
  year={2021},
  publisher={Elsevier}
}

@inproceedings{chen2014flu,
  title={Flu gone viral: Syndromic surveillance of flu on twitter using temporal topic models},
  author={Chen, Liangzhe and Hossain, KSM Tozammel and Butler, Patrick and Ramakrishnan, Naren and Prakash, B Aditya},
  booktitle={2014 IEEE international conference on data mining},
  pages={755--760},
  year={2014},
  organization={IEEE}
}

@article{ginsberg2009detecting,
  title={Detecting influenza epidemics using search engine query data},
  author={Ginsberg, Jeremy and Mohebbi, Matthew H and Patel, Rajan S and Brammer, Lynnette and Smolinski, Mark S and Brilliant, Larry},
  journal={Nature},
  volume={457},
  number={7232},
  pages={1012--1014},
  year={2009},
  publisher={Nature Publishing Group UK London}
}

@article{mciver2014wikipedia,
  title={Wikipedia usage estimates prevalence of influenza-like illness in the United States in near real-time},
  author={McIver, David J and Brownstein, John S},
  journal={PLoS computational biology},
  volume={10},
  number={4},
  pages={e1003581},
  year={2014},
  publisher={Public Library of Science San Francisco, USA}
}

@article{polgreen2007use,
  title={Use of prediction markets to forecast infectious disease activity},
  author={Polgreen, Philip M and Nelson, Forrest D and Neumann, George R and Weinstein, Robert A},
  journal={Clinical Infectious Diseases},
  volume={44},
  number={2},
  pages={272--279},
  year={2007},
  publisher={The University of Chicago Press}
}

@article{smolinski2015flu,
  title={Flu near you: crowdsourced symptom reporting spanning 2 influenza seasons},
  author={Smolinski, Mark S and Crawley, Adam W and Baltrusaitis, Kristin and Chunara, Rumi and Olsen, Jennifer M and W{\'o}jcik, Oktawia and Santillana, Mauricio and Nguyen, Andre and Brownstein, John S},
  journal={American journal of public health},
  volume={105},
  number={10},
  pages={2124--2130},
  year={2015},
  publisher={American Public Health Association}
}

@article{salomon2021us,
  title={The US COVID-19 Trends and Impact Survey: Continuous real-time measurement of COVID-19 symptoms, risks, protective behaviors, testing, and vaccination},
  author={Salomon, Joshua A and Reinhart, Alex and Bilinski, Alyssa and Chua, Eu Jing and La Motte-Kerr, Wichada and R{\"o}nn, Minttu M and Reitsma, Marissa B and Morris, Katherine A and LaRocca, Sarah and Farag, Tamer H and others},
  journal={Proceedings of the National Academy of Sciences},
  volume={118},
  number={51},
  pages={e2111454118},
  year={2021},
  publisher={National Acad Sciences}
}

@article{miller2018smartphone,
  title={A smartphone-driven thermometer application for real-time population-and individual-level influenza surveillance},
  author={Miller, Aaron C and Singh, Inder and Koehler, Erin and Polgreen, Philip M},
  journal={Clinical Infectious Diseases},
  volume={67},
  number={3},
  pages={388--397},
  year={2018},
  publisher={Oxford University Press US}
}

@article{leuba2020tracking,
  title={Tracking and predicting US influenza activity with a real-time surveillance network},
  author={Leuba, Sequoia I and Yaesoubi, Reza and Antillon, Marina and Cohen, Ted and Zimmer, Christoph},
  journal={PLoS computational biology},
  volume={16},
  number={11},
  pages={e1008180},
  year={2020},
  publisher={Public Library of Science San Francisco, CA USA}
}

@article{nsoesie2014guess,
  title={Guess who’s not coming to dinner? Evaluating online restaurant reservations for disease surveillance},
  author={Nsoesie, Elaine O and Buckeridge, David L and Brownstein, John S},
  journal={Journal of medical Internet research},
  volume={16},
  number={1},
  pages={e2998},
  year={2014},
  publisher={JMIR Publications Inc., Toronto, Canada}
}

@article{miliou2021predicting,
  title={Predicting seasonal influenza using supermarket retail records},
  author={Miliou, Ioanna and Xiong, Xinyue and Rinzivillo, Salvatore and Zhang, Qian and Rossetti, Giulio and Giannotti, Fosca and Pedreschi, Dino and Vespignani, Alessandro},
  journal={PLOS Computational Biology},
  volume={17},
  number={7},
  pages={e1009087},
  year={2021},
  publisher={Public Library of Science San Francisco, CA USA}
}

@article{pepe2020covid,
  title={COVID-19 outbreak response, a dataset to assess mobility changes in Italy following national lockdown},
  author={Pepe, Emanuele and Bajardi, Paolo and Gauvin, Laetitia and Privitera, Filippo and Lake, Brennan and Cattuto, Ciro and Tizzoni, Michele},
  journal={Scientific data},
  volume={7},
  number={1},
  pages={230},
  year={2020},
  publisher={Nature Publishing Group UK London}
}

@article{klise2021analysis,
  title={Analysis of mobility data to build contact networks for COVID-19},
  author={Klise, Katherine and Beyeler, Walt and Finley, Patrick and Makvandi, Monear},
  journal={PLoS One},
  volume={16},
  number={4},
  pages={e0249726},
  year={2021},
  publisher={Public Library of Science San Francisco, CA USA}
}

@inproceedings{mulay2020pandemic,
  title={Pandemic spread prediction and healthcare preparedness through financial and mobility data},
  author={Mulay, Nidhi and Bishnoi, Vikas and Charotia, Himanshi and Asthana, Siddhartha and Dhama, Gaurav and Arora, Ankur},
  booktitle={2020 19th IEEE International Conference on Machine Learning and Applications (ICMLA)},
  pages={1340--1347},
  year={2020},
  organization={IEEE}
}

@article{chan:chb21,
author = {Chan, Eugene and Saqib, Najam},
year = {2021},
month = {01},
pages = {106718},
title = {Privacy Concerns Can Explain Unwillingness to Download and Use Contact Tracing Apps When COVID-19 Concerns are High},
volume = {119},
journal = {Computers in Human Behavior},
doi = {10.1016/j.chb.2021.106718}
}

@article{tran2021health,
  title={Health vs. privacy? The risk-risk tradeoff in using COVID-19 contact-tracing apps},
  author={Tran, Cong Duc and Nguyen, Tin Trung},
  journal={Technology in Society},
  volume={67},
  pages={101755},
  year={2021},
  publisher={Elsevier}
}

@inproceedings{li2024computing,
  title={Computing epidemic metrics with edge differential privacy},
  author={Li, George Z and Nguyen, Dung and Vullikanti, Anil},
  booktitle={International Conference on Artificial Intelligence and Statistics},
  pages={4303--4311},
  year={2024},
  organization={PMLR}
}

@article{li2022differentially,
  title={Differentially private partial set cover with applications to facility location},
  author={Li, George Z and Nguyen, Dung and Vullikanti, Anil},
  journal={IJCAI},
  year={2022}
}

@inproceedings{adiga2022enhancing,
  title={Enhancing COVID-19 ensemble forecasting model performance using auxiliary data sources},
  author={Adiga, Aniruddha and Kaur, Gursharn and Hurt, Benjamin and Wang, Lijing and Porebski, Przemyslaw and Venkatramanan, Srinivasan and Lewis, Bryan and Marathe, Madhav},
  booktitle={2022 IEEE International Conference on Big Data (Big Data)},
  pages={1594--1603},
  year={2022},
  organization={IEEE}
}

@article{adiga2020mathematical,
  title={Mathematical models for covid-19 pandemic: a comparative analysis},
  author={Adiga, Aniruddha and Dubhashi, Devdatt and Lewis, Bryan and Marathe, Madhav and Venkatramanan, Srinivasan and Vullikanti, Anil},
  journal={Journal of the Indian Institute of Science},
  volume={100},
  number={4},
  pages={793--807},
  year={2020},
  publisher={Springer}
}

@article{xu2020epidemiological,
  title={Epidemiological data from the COVID-19 outbreak, real-time case information},
  author={Xu, Bo and Gutierrez, Bernardo and Mekaru, Sumiko and Sewalk, Kara and Goodwin, Lauren and Loskill, Alyssa and Cohn, Emily L and Hswen, Yulin and Hill, Sarah C and Cobo, Maria M and others},
  journal={Scientific data},
  volume={7},
  number={1},
  pages={1--6},
  year={2020},
  publisher={Nature Publishing Group}
}

@misc{COVID19C29:online,
author = {CDC},
title = {COVID-19 Case Surveillance Public Use Data | Data | Centers for Disease Control and Prevention},
howpublished = {\url{https://data.cdc.gov/Case-Surveillance/COVID-19-Case-Surveillance-Public-Use-Data/vbim-akqf}},
month = {},
year = {},
note = {(Accessed on 08/24/2020)}
}

@article{adiga2020data,
  title={Data-driven modeling for different stages of pandemic response},
  author={Adiga, Aniruddha and Chen, Jiangzhuo and Marathe, Madhav and Mortveit, Henning and Venkatramanan, Srinivasan and Vullikanti, Anil},
  journal={Journal of the Indian Institute of Science},
  volume={100},
  number={4},
  pages={901--915},
  year={2020},
  publisher={Springer}
}

@techreport{abowd20232010,
  title={The 2010 Census Confidentiality Protections Failed, Here's How and Why},
  author={Abowd, John M and Adams, Tamara and Ashmead, Robert and Darais, David and Dey, Sourya and Garfinkel, Simson L and Goldschlag, Nathan and Kifer, Daniel and Leclerc, Philip and Lew, Ethan and others},
  year={2023},
  institution={National Bureau of Economic Research}
}

@inproceedings{abowd2018us,
  title={{The US Census Bureau adopts differential privacy}},
  author={Abowd, John M},
  booktitle={Proceedings of the 24th ACM SIGKDD international conference on knowledge discovery \& data mining},
  pages={2867--2867},
  year={2018}
}

@article{analyticsexposure,
  title={Exposure Notification Privacy-Preserving Analytics (ENPA) White Paper},
  author={Analytics, Exposure Notification Privacy-Preserving},
  journal={ENPA White Paper.pdf}
}

@article{team2017learning,
  title={Learning with privacy at scale},
  author={Team, ADP and others},
  journal={Apple Mach. Learn. J},
  volume={1},
  number={8},
  pages={1--25},
  year={2017}
}

@inproceedings{wu2022linkteller,
  title={Linkteller: Recovering private edges from graph neural networks via influence analysis},
  author={Wu, Fan and Long, Yunhui and Zhang, Ce and Li, Bo},
  booktitle={2022 ieee symposium on security and privacy (sp)},
  pages={2005--2024},
  year={2022},
  organization={IEEE}
}

@article{gong2018attribute,
  title={Attribute inference attacks in online social networks},
  author={Gong, Neil Zhenqiang and Liu, Bin},
  journal={ACM Transactions on Privacy and Security (TOPS)},
  volume={21},
  number={1},
  pages={1--30},
  year={2018},
  publisher={ACM New York, NY, USA}
}

@inproceedings{zari2024node,
  title={Node injection link stealing attack},
  author={Zari, Oualid and Parra-Arnau, Javier and {\"U}nsal, Ay{\c{s}}e and {\"O}nen, Melek},
  booktitle={International Conference on Privacy in Statistical Databases},
  pages={358--373},
  year={2024},
  organization={Springer}
}

@article{ponomareva2023dp,
  title={How to dp-fy ml: A practical guide to machine learning with differential privacy},
  author={Ponomareva, Natalia and Hazimeh, Hussein and Kurakin, Alex and Xu, Zheng and Denison, Carson and McMahan, H Brendan and Vassilvitskii, Sergei and Chien, Steve and Thakurta, Abhradeep Guha},
  journal={Journal of Artificial Intelligence Research},
  volume={77},
  pages={1113--1201},
  year={2023}
}

@article{ji2014differential,
  title={Differential privacy and machine learning: a survey and review},
  author={Ji, Zhanglong and Lipton, Zachary C and Elkan, Charles},
  journal={arXiv preprint arXiv:1412.7584},
  year={2014}
}

@article{blanco2022critical,
  title={A critical review on the use (and misuse) of differential privacy in machine learning},
  author={Blanco-Justicia, Alberto and S{\'a}nchez, David and Domingo-Ferrer, Josep and Muralidhar, Krishnamurty},
  journal={ACM Computing Surveys},
  volume={55},
  number={8},
  pages={1--16},
  year={2022},
  publisher={ACM New York, NY}
}

@inproceedings{venkatramanan2017spatio,
  title={Spatio-temporal optimization of seasonal vaccination using a metapopulation model of influenza},
  author={Venkatramanan, Srinivasan and Chen, Jiangzhuo and Gupta, Sandeep and Lewis, Bryan and Marathe, Madhav and Mortveit, Henning and Vullikanti, Anil},
  booktitle={2017 IEEE International Conference on Healthcare Informatics (ICHI)},
  pages={134--143},
  year={2017},
  organization={IEEE}
}

@inproceedings{zhang2020doubleensemble,
  title={DoubleEnsemble: A new ensemble method based on sample reweighting and feature selection for financial data analysis},
  author={Zhang, Chuheng and Li, Yuanqi and Chen, Xi and Jin, Yifei and Tang, Pingzhong and Li, Jian},
  booktitle={2020 IEEE international conference on data mining (ICDM)},
  pages={781--790},
  year={2020},
  organization={IEEE}
}

@article{zhang2022differentially,
  title={Differentially private real-time release of sequential data},
  author={Zhang, Xueru and Khalili, Mohammad Mahdi and Liu, Mingyan},
  journal={ACM Transactions on Privacy and Security},
  volume={26},
  number={1},
  pages={1--29},
  year={2022},
  publisher={ACM New York, NY}
}

@article{ghazi2023user,
  title={User-level differential privacy with few examples per user},
  author={Ghazi, Badih and Kamath, Pritish and Kumar, Ravi and Manurangsi, Pasin and Meka, Raghu and Zhang, Chiyuan},
  journal={Advances in Neural Information Processing Systems},
  volume={36},
  pages={19263--19290},
  year={2023}
}

@inproceedings{dwork2010differential,
  title={Differential privacy under continual observation},
  author={Dwork, Cynthia and Naor, Moni and Pitassi, Toniann and Rothblum, Guy N},
  booktitle={Proceedings of the forty-second ACM symposium on Theory of computing},
  pages={715--724},
  year={2010}
}

@article{choi2012predicting,
  title={Predicting the present with Google Trends},
  author={Choi, Hyunyoung and Varian, Hal},
  journal={Economic record},
  volume={88},
  pages={2--9},
  year={2012},
  publisher={Wiley Online Library}
}

@article{gunther2021nowcasting,
  title={Nowcasting the COVID-19 pandemic in Bavaria},
  author={G{\"u}nther, Felix and Bender, Andreas and Katz, Katharina and K{\"u}chenhoff, Helmut and H{\"o}hle, Michael},
  journal={Biometrical Journal},
  volume={63},
  number={3},
  pages={490--502},
  year={2021},
  publisher={Wiley Online Library}
}

@article{quevedo2024unveiling,
  title={Unveiling pandemic patterns: a detailed analysis of transmission and severity parameters across four COVID-19 waves in Bogot{\'a}, Colombia},
  author={Quevedo, David Santiago and Dom{\'\i}nguez, Nicol{\'a}s T and Perez, Diego Fernando and Cabrera Polan{\'\i}a, Maria Alejandra and Serrano Medina, Juan David and Abril-Berm{\'u}dez, Felipe Segundo and Romero, Diane Moyano and Rios Oliveros, Diana Sofia and Gonz{\'a}lez Mayorga, Manuel Alfredo and Whittaker, Charles and others},
  journal={BMC Global and Public Health},
  volume={2},
  number={1},
  pages={83},
  year={2024},
  publisher={Springer}
}

@article{Cori2013,
 author={Cori, A and Ferguson, NM and Fraser, C and Cauchemez, S},
 year={2013},
 title={{A New Framework and Software to Estimate Time-Varying Reproduction Numbers During Epidemics}},
 journal={Am. J. Epidemiol.},
 doi={10.1093/aje/kwt133},
}

\end{document}